\documentclass[Afour,sageh,times]{sagej}

\usepackage{moreverb,url}

\usepackage[colorlinks,bookmarksopen,bookmarksnumbered,citecolor=red,urlcolor=red]{hyperref}
\usepackage{subcaption}
\usepackage{acronym}
\usepackage{amsmath, amssymb, amsthm}
\usepackage{algorithm, algpseudocode}
\usepackage{mathtools}
\usepackage{siunitx}
\usepackage{ulem}
\usepackage{bm}
\usepackage{nicefrac}
\usepackage{tikz}
\usetikzlibrary{positioning}

\newcommand\BibTeX{{\rmfamily B\kern-.05em \textsc{i\kern-.025em b}\kern-.08em
T\kern-.1667em\lower.7ex\hbox{E}\kern-.125emX}}

\def\volumeyear{2016}

\acrodef{RL}{Reinforcement Learning}
\acrodef{PPO}{Proximal Policy Optimization}
\acrodef{MPC}{Model Predictive Control}
\acrodef{ILC}{Iterative Learning Control}
\acrodef{NN}{Neural Network}
\acrodef{SGD}{Stochastic Gradient Descent}
\acrodef{PL}{Polyak-{\L}ojasiewicz}
\acrodef{BPTT}{Back Propagation Through Time}
\acrodef{MPPI}{Model Predictive Path Integral}
\acrodef{FEM}{Finite Element Method}
\acrodef{MSE}{Mean Squared Error}
\acrodef{DDPG}{Deep Deterministic Policy Gradient}
\acrodef{TV}{Total Variation}

\usepackage{xcolor}

\newcommand{\revision}[1]{\textcolor[RGB]{0, 0, 0}{#1}}

\begin{document}

\runninghead{Nan et al.}

\title{Efficient Model-Based Reinforcement Learning for Robot Control via Online \revision{Optimization}}

\author{
Fang Nan\affilnum{1},
Hao Ma\affilnum{2, 3},
Qinghua Guan\affilnum{4},
Josie Hughes\affilnum{4},
Michael Muehlebach\affilnum{2},
and Marco Hutter\affilnum{1}}

\affiliation{\affilnum{1}Robotic Systems Lab, ETH Z{\"u}rich, Z{\"u}rich, Switzerland\\
\affilnum{2}Learning and Dynamical Systems, Max Planck Institute for Intelligent Systems, T{\"u}bingen, Germany\\
\affilnum{3}Institute for Dynamic Systems and Control, ETH Z{\"u}rich, Z{\"u}rich, Switzerland\\
\affilnum{4}CREATE Lab, EPFL, Lausanne, Switzerland}

\corrauth{Fang Nan, Robotic Systems Lab, ETH Z{\"u}rich, Leonhardstrasse 21, 8092 Z{\"u}rich, Switzerland.}
\email{fannan@ethz.ch}

\begin{abstract}
We present an online model-based reinforcement learning algorithm suitable for controlling complex robotic systems directly in the real world.
Unlike prevailing sim-to-real pipelines that rely on extensive offline simulation and model-free policy optimization, our method builds a dynamics model from real-time interaction data and performs policy updates guided by the learned dynamics model.
This efficient model-based reinforcement learning scheme significantly reduces the number of samples to train control policies, enabling direct training on real-world rollout data.
This significantly reduces the influence of bias in the simulated data, and facilitates the search for high-performance control policies.
We adopt online \revision{optimization} analysis to derive sublinear regret bounds under stochastic online optimization assumptions, providing formal guarantees on performance improvement as more interaction data are collected.
Experimental evaluations were performed on a hydraulic excavator arm and a soft robot arm, where the algorithm demonstrates strong sample efficiency compared to model-free reinforcement learning methods, reaching comparable performance within hours.
Robust adaptation to shifting dynamics was also observed when the payload condition was randomized.
Our approach paves the way toward efficient and reliable on-robot learning for a broad class of challenging control tasks.
\end{abstract}


\maketitle
\section{Introduction}
\label{sec:introduction}


In recent years, \ac{RL} has achieved remarkable success across a variety of robotic tasks, ranging from manipulation~\citep{Kroemer21ReviewRobot} and locomotion~\cite{Miki22LearningRobust} to aerial vehicle control~\citep{Kaufmann23ChampionlevelDrone}.
By leveraging neural networks and end-to-end learning algorithms, agents can acquire policies that directly map high-dimensional sensor inputs to low-level motor commands~\citep{Lin25SimtoRealReinforcement}.

The prevailing \ac{RL} workflow in robotics today typically follows a sim-to-real paradigm.
In this pipeline, control policies are first trained extensively in physical simulators to gather large amounts of data.
After achieving satisfying performance in simulation, the policies are transferred to physical robots and used in real-world applications.
This approach has been widely adopted due to its ability to leverage the efficiency of simulation for data collection, allowing for rapid iteration and exploration of complex environments.
In particular, popular model-free \ac{RL} algorithms, such as \ac{PPO}~\citep{Schulman17ProximalPolicy}, require large amounts of data to effectively train control policies, making data collection in simulation an indispensable step.

This sim-to-real pipeline, however, comes with significant limitations.
First, the fidelity of simulators is often insufficient to capture the nuanced dynamics of complex robotic hardware, leading to a sim-to-real gap that degrades performance when policies are deployed in the real world.
In some cases, researchers have reported that \ac{RL} algorithms exploit the simulator's imperfections and learn policies that cannot be transferred to the real world~\citep{Kadian20Sim2RealPredictivity}.
A number of strategies have been proposed to mitigate this issue, such as domain randomization~\citep{Tobin17DomainRandomization}, real-to-sim~\citep{Villasevil24ReconcilingReality}, and domain adaptation~\citep{Kumar21RMARapid}.
While these methods can improve the transferability of policies, extensive tuning effort and a careful tradeoff between performance and robustness are often required.

The applicability of the sim-to-real paradigm is further limited by the lack of simulators for complex robotic systems.
Robotic platforms with hydraulic or pneumatic actuators, deformable soft structures, or robots that interact with non-rigid objects have distinct applications and are vital for many real-world tasks.
Existing simulators for robot learning, while being very efficient in simulating rigid-body systems with torque-controlled joints, have minimal support for these complex systems~\citep{Makoviychuk21IsaacGym}.
Although specialized simulators exist for such multidomain physical simulation~\citep{TheMathWorksInc.25SimscapeMultidomain, AlgoryxSimulationAB25AGXDynamics}, their efficiency is not sufficient for parallelized data collection in robot learning applications.

These limitations of the sim-to-real paradigm motivate alternative \ac{RL} algorithms that can operate directly on physical robots, and model-based \ac{RL} methods have therefore emerged as a promising approach.
Unlike model-free methods, model-based \ac{RL} algorithms learn a dynamics model of the environment and use the model to plan actions or improve a policy.
As the model learning can be performed with off-policy data, these methods can significantly reduce the amount of real-world data required for training.
Existing model-based \ac{RL} algorithms, however, have disadvantages in terms of algorithm efficiency, making them hard to deploy in real-world learning settings~\citep{Romero25DreamFly}.

\subsection{Contributions}
To address the limitations mentioned above, we propose a novel model-based \ac{RL} algorithm designed for online training directly on physical robots.
Our approach constructs a dynamics model from real-time interaction data and uses the learned dynamics model to generate approximate gradients for policy optimization on real-world data.
This paradigm significantly reduces the need for generating a large amount of synthetic data during policy optimization or online planning, thus achieving faster convergence and lower computational cost.
Moreover, by operating entirely on data collected on the real system, our method gracefully handles unmodeled phenomena, thus extending applicability to robots with complex or partially unknown dynamics.

We support the proposed approach with theoretical analysis from the perspective of stochastic online optimization.
In particular, we analyze regret of the online model and policy learning modules in the proposed method under some simplifications and discuss necessary conditions for joint convergence.

Finally, we demonstrate the efficacy of our method through experiments on two robotic platforms: a hydraulic excavator arm (HEAP) and a modular soft robot arm.
The real-world experiments show that our method achieves strong sample efficiency toward learning control policies, enabling \ac{RL} directly on physical robots.

The remainder of the paper is organized as follows.
Section 2 reviews related work in model-based and model-free RL, with an emphasis on their application to robot control problems.
Section 3 formalizes the problem setting, introduces our \revision{model-based \ac{RL} algorithm with online optimization}, and presents the theoretical analysis leading to regret bounds.
Section 4 details our experimental setup on multiple robotic platforms.
Section 5 reports empirical results, comparing against state-of-the-art baselines.
Finally, Section 6 concludes with a discussion of \revision{ablation studies}, limitations and future directions.

\section{Related Work}
\label{sec:related_work}

\subsection{Model-based RL in Robotics}
\ac{RL} has shown great performance in various robotic applications.
In particular, model-free \ac{RL} has become the dominant approach for training control policies for legged locomotion~\citep{Miki22LearningRobust}.
Other successful examples include racing drones~\citep{Kaufmann23ChampionlevelDrone}, robotic manipulation~\citep{Kroemer21ReviewRobot}, and control of agile robotic systems~\citep{Ma23ReinforcementLearning, Buchler22LearningPlay}.
These applications typically use model-free policy optimization algorithms, such as \ac{PPO}~\citep{Schulman17ProximalPolicy}, to learn control policies from high-dimensional sensory inputs.
To achieve this, the control policies are often parameterized by deep neural networks.
Policy optimization algorithms use massive rollout data to estimate the local gradient of the performance objective and then update the policy parameters in the direction of the estimated gradient.
Thus, policy optimization algorithms are often on-policy and require large amounts of data to learn effective policies.

A promising alternative to model-free \ac{RL} is model-based \ac{RL}.
In the context of robotics, this refers to algorithms that learn a dynamics model of the environment and use the model to plan actions or improve a policy.
As model learning can be performed with off-policy data, these methods can significantly reduce the amount of real-world data required for training.
The learned models are often used in two ways: (1) to generate synthetic data for policy optimization, and (2) to perform online planning.

\textit{Data generation for policy optimization:} The first use case dates back to the early days of \ac{RL}, when the Dyna approach was proposed~\citep{Sutton90IntegratedModeling}.
In this framework, a model of the environment is learned from real-world data and then used to generate synthetic data.
Model-free \ac{RL} algorithms are then applied on both real-world and synthetic data to optimize the policy.
The approach originally showed effective at solving tabular \ac{RL} problems, and has been extended to deep \ac{RL}.
The approach by~\citet{Ha18RecurrentWorld} handles image inputs that are encoded to a latent space where they are subject to a latent dynamics model.
The idea was continued in the Dreamer algorithms~\citep{Hafner19DreamControl, Hafner20MasteringAtari, Hafner25MasteringDiverse}, which use learned latent dynamics models to generate synthetic data for policy training with actor-critic algorithms.
The series of Dreamer algorithms has evolved over the years, and the latest version, DreamerV3~\citep{Hafner25MasteringDiverse}, has shown the capability of learning complex behaviors in computer games and the possibility of handling different tasks with the same set of hyperparameters.
There have been a few works that successfully applied the Dreamer algorithms to real-world robot control tasks, such as quadrupedal locomotion~\citep{Wu23DayDreamerWorld,Li25RoboticWorld,Li25OfflineRobotic}, solving a labyrinth game~\citep{Bi24SampleEfficientLearning}, and vision-based quadcopter control~\citep{Romero25DreamFly}.
In all of these works, advantages in sample efficiency have been reported.
However, it was also noticed that running the model-free \ac{RL} algorithms repeatedly with simulated data leads to the algorithm being the bottleneck of the training process instead of the data collection~\citep{Romero25DreamFly}.
Consequently, some existing works still perform model-based \ac{RL} in simulation before transferring to the real world, implying they may still be affected by the sim-to-real gap~\citep{Li25RoboticWorld, Romero25DreamFly}.

\textit{Online planning via learned model:} The second use-case takes advantage of the learned model for online planning.
Some early works in this direction perform model predictive control with learned models to provide expert samples for model-free \ac{RL} finetuning~\citep{Nagabandi18NeuralNetwork}.
A prominent example is the TD-MPC algorithm~\citep{Hansen22TemporalDifference, Hansen24TDMPC2Scalable}, which combines model learning with online planning.
In this approach, a latent-space dynamics model is learned from real-world data, along with a terminal value function.
The learned models are subsequently used to perform online planning with a model predictive path integral search.
This approach achieves great sample efficiency by leveraging off-policy data across multiple episodes for model learning.
However, a significant computational overhead arises from the online sampling-based planning.

\subsection{Learning-based Control for Complex Robot Systems}
Although a wide range of successful applications of \ac{RL} have been reported in the literature, many are limited to rigid-body systems with torque-controlled joints.
In contrast, a wide range of real-world robotic systems are equipped with hydraulic or pneumatic actuators, or involve non-rigid objects as part of the robot's body or the environment.
The control of these systems is often challenging, in particular due to the lack of simulators that match the accuracy and computational efficiency of their rigid-body counterparts.
A number of methods have been proposed to address the lack of simulation tools.
We present some representative examples in the context of hydraulic robots and soft robots, which are the platforms used to evaluate our approach.

The control of hydraulic actuators in robots is often considered in the context of the automation of heavy machinery.
A number of works considered building a physics-based model of the hydraulic system, and then using nonlinear model-based control methods to develop controllers~\citep{Mattila17SurveyControl}.
These methods, however, often face challenges when applied to real-world hydraulic systems, where the system dynamics cannot be fully understood.
In this case, data-driven modeling has emerged as a promising alternative.
Some early works use system identification methods to learn parameters in a physics-based model~\citep{Nurmi17AutomatedFeedForward, Nurmi18NeuralNetwork}, which is then plugged into a model-based control framework.
Later research used data-driven modeling to learn the complete dynamics model of the hydraulic system, and then used the model as a simulation environment for model-free \ac{RL}~\citep{Egli22GeneralApproach}.
As the learned model contains the complex dynamics and coupling effects of the hydraulic system, the corresponding controller is able to achieve high-performance control on the modeled system.
Subsequent works have combined physical modeling with learning-based methods to learn the residual dynamics of the hydraulic system, achieving both high performance and training efficiency~\citep{Lee22PrecisionMotion}.
These \ac{RL}-based methods, however, are limited by simulation environments.
The trained policies only work well on the target system where the model is trained, and the performance degrades if the model is biased.
This issue is partially addressed by the online adaptation approach proposed in~\citep{Nan24LearningAdaptive}, where the control policy is trained in simulation with a set of randomized system parameters and then adapted to the real-world system rapidly at the time of deployment.
Real-world experiments show that this approach is able to achieve high performance control on multiple hydraulic systems using online adaptation.
The method, however, still requires a good understanding of the system model and careful selection of the parameters to be randomized, which limits its generalization to distinctly different systems.

Robots with soft structures share a lot of similar challenges in modeling and control.
Modeling the dynamics of soft links with finite elements is accurate but slow for control purposes, while simplified models are often not sufficient to achieve performant control; thus, data-driven modeling is often recommended~\citep{Faure12SOFAMultiModel, Spinelli22UnifiedModular}.
Previously, a number of works have used deep learning to predict the dynamics of soft robots for data synthesis and predictive control purposes~\citep{Chen25DataDrivenMethods, Sapai23DeepLearning, Chen24HybridAdaptive}.
In~\citep{Li22AdaptiveContinuous}, a framework for an \ac{RL} method for a pneumatically actuated soft robot is proposed.
The approach combines pre-training in simulation and fine-tuning on real-world data in \ac{RL} with Deep Deterministic Policy Gradient to efficiently learn an accurate controller for a three-chamber soft arm.
The proposed method, however, still relies on an approximately accurate simulation model, and the real-world training is time-consuming.
Researchers have also shown the possibility of performing optimization-based control with learned dynamics models~\citep{Bern20SoftRobot}.
However, as pointed out by the authors, the approach is dependent on the accuracy of the learned model and does not have any capability for online refinement.


\subsection{Online Learning for Control}
Online learning provides a framework for sequential decision-making, where an agent iteratively improves its decisions based on feedback from the environment~\citep{Hazan19IntroductionOnline}.
The potential of integrating online learning principles into \ac{RL} and control frameworks has recently emerged as a promising direction to achieve adaptive and robust control~\citep{Hazan25IntroductionOnline}.

Recent work by~\citet{Ma25StochasticOnline} draws connections between online learning and policy optimization.
Extending the classic approach of iterative learning control~\citep{Hofer19IterativeLearning, Ma22LearningbasedIterative}, they embed an approximate dynamics model into a stochastic, gradient-based online optimization loop.
Their algorithm unifies first-order gradient descent and quasi-Newton updates using model-informed gradient estimates.
The connection between modeling error and convergence is quantified, and sublinear regret is proven under general smoothness and bounded-error conditions.
Experiments on a flexible beam, a quadruped robot, and a ping-pong robot demonstrate robust, real-time adaptation, even in the presence of noisy gradients.
However, this approach focuses mainly on feedforward controller optimization and assumes access to a known model with bounded error.

\citet{Lin24OnlinePolicy} addresses the limitation of requiring a known model by introducing a meta-framework that decouples the online policy optimizer (ALG) from the estimator of unknown dynamics (EST).
They show that if ALG, when given perfect dynamics, satisfies certain contractivity and stability properties, then coupling ALG with EST only introduces bounded perturbations.
The cumulative effect of these perturbations scales with local estimation errors along the visited trajectory.

In this paper, we extend these strands by proposing a model-based \ac{RL} algorithm with online learning inspirations.
The algorithm includes an online model learning module to learn an explicit dynamics model, and uses its gradients to guide control policy updates.
We provide a theoretical analysis to explain the effectiveness of the proposed online learning algorithm in nonlinear robot control problems, and demonstrate the effectiveness of the approach through real-world robot experiments.

\section{Methodology}
\label{sec:methodology}


\subsection{Problem Formulation}
We consider the problem of controlling an unknown dynamical system by minimizing a cost function over time.
The system is described by a state $x_{\tau} \in  \mathcal{X} \subseteq \mathbb{R}^n$ at each time step $\tau$
, and the control input is denoted by $u_{\tau} \in \mathcal{U} \subseteq \mathbb{R}^m$. We note that both $\mathcal{X}$ and $\mathcal{U}$ are bounded and convex. The system evolves according to the dynamics:
\begin{equation}
  x_{\tau+1} = f(x_{\tau}, u_{\tau}),
  \label{eq:def_dynamics}
\end{equation}
where $f: \mathcal{X} \times \mathcal{U} \to \mathcal{X}$ is a function describing the system dynamics.
\revision{
Given a sequence of references or commands $\tilde{x}_{0}, \tilde{x}_{1}, \ldots \in \widetilde{\mathcal{X}} \subseteq \mathbb{R}^{p}$, we aim to find a control policy $\pi: \mathcal{X} \times \widetilde{\mathcal{X}} \to \mathcal{U}$, to minimize the expected cost in an episodic setting:
}
\begin{align}
\begin{split}
      \min \ &~\mathbb{E} \left[ \sum_{t=0}^\infty \sum_{\tau=0}^{H-1} c(x_\tau^{(t)}, u_\tau^{(t)}, \tilde{x}_\tau^{(t)}) \right]\\
  \text{s.t. } &~x_{\tau+1}^{(t)} = f(x_\tau^{(t)}, u_\tau^{(t)})\\
  &~u_\tau^{(t)} = \pi(x_\tau^{(t)}, \tilde{x}_\tau^{(t)}),
  ~\forall t\geq0,~\tau\geq0
\end{split}
\label{eq:J_of_pi}
\end{align}
where $x_\tau^{(t)}, \tau = 0, \dots, H-1$ is the state evolution according to~\eqref{eq:def_dynamics} over episode $t$ of length $H$.
The cost function is denoted by \revision{$c: \mathcal{X} \times \mathcal{U} \times \tilde{\mathcal{X}} \to \mathbb{R}$}, and may depend on the state, the control input, and an auxiliary random variable $\tilde{x}_{\tau} \in \widetilde{\mathcal{X}}$, which represents reference trajectories or user commands.
The reference trajectories $\tilde{\bm{x}}_t = \left(\tilde{x}_{0}^{(t)}, \dots, \tilde{x}_{H-1}^{(t)}\right) \overset{\text{i.i.d.}}{\sim} \widetilde{\mathcal{D}}$ are sampled at the beginning at each episode $t$, and $x_0^{(t)}$ is initialized according to $\tilde{x}_0^{(t)}$.
\revision{
While we consider training in an episodic setting with a fixed horizon $H$, the learned control policy is not limited to tasks of the same length. Since the episodes are sufficiently long and the reference trajectories are randomized, the policy, which operates on a local window of states and references, can generalize to reference sequences of different lengths.
}

\subsection{Algorithm}
\begin{figure}
  \centering
  \includegraphics[width=\linewidth]{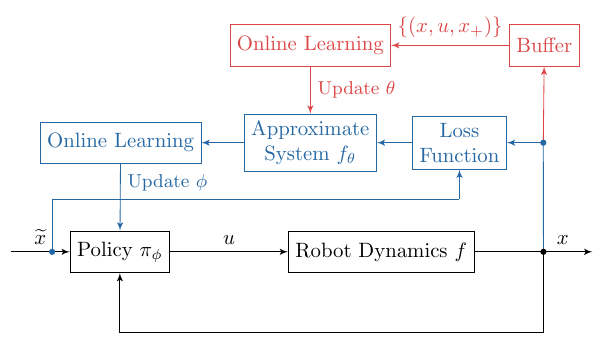}
  \caption{A diagram showing the workflow of the proposed online model-based \ac{RL} algorithm. It involves online model learning and policy optimization at the same time. The red blocks highlight the process of model learning, and the blue components show the workflow of policy optimization.}
  \label{fig:online_mbrl_diagram}
\end{figure}

To address~\eqref{eq:J_of_pi} we propose a model-based \ac{RL} algorithm that leverages the estimated system dynamics to optimize the control policy. We use a \ac{NN} $f_\theta$ to model the true dynamics function $f$, and the deterministic control policy is represented by another \ac{NN} $\pi_\phi$.
The parameters $\theta \in \Theta \subseteq \mathbb{R}^{n_{\theta}}$ and $\phi\in \Phi \subseteq \mathbb{R}^{n_{\phi}}$ are optimized using stochastic gradient descent in an online manner.
We also note that both $\Theta$ and $\Phi$ are bounded and convex.
The proposed online model-based RL algorithm is outlined in Algorithm~\ref{alg:online_mbrl}.
Building on top of previous work on approximate gradient-based online \revision{optimization}, Algorithm~\ref{alg:online_mbrl} simultaneously learns a dynamics model online for calculating approximate policy gradients instead of assuming access to a known model.
A diagram illustrating the workflow of the proposed algorithm is shown in Fig.~\ref{fig:online_mbrl_diagram}.

\begin{algorithm}[thb]
  \caption{Online Model-Based RL}
  \label{alg:online_mbrl}
  \begin{algorithmic}[1]
    \State Initialize parameters $\theta$, $\phi$ and let $\mathcal D_0 = \emptyset$
      \For{$t = 0, 1, \ldots, T-1$}
        \State Sample reference $\tilde{x}_{0:H}^{(t)} \in \widetilde{\mathcal{D}}$
        \State Initialize state $x_0 = \tilde{x}_0^{(t)}$
        \For {$\tau = 0, 1, \ldots, H$}
          \State Compute control input $u_\tau^{(t)} = \pi_\phi(x_\tau^{(t)}, \tilde{x}_\tau^{(t)})$
          \State Apply $u_\tau^{(t)}$ and observe next state $x_{\tau+1}^{(t)}$
        \EndFor
        \State Extend the buffer:
        \State \quad $\mathcal D_{t+1} \gets \mathcal D_{t} \cup \{(x_{\tau}^{(t)}, u_{\tau}^{(t)}, x_{\tau+1}^{(t)})\}_{\tau=0}^{H-1}$
        \State Update dynamics model: 
        \State \quad $\theta_{t+1} \gets \theta_t - \alpha_\theta \nabla_\theta \ell(\theta)$
        \Statex \hfill \Comment{Using~\eqref{eq:model_learning_loss} evaluated on mini-batchs from $\mathcal D_{t+1}$}
        \State Calculate controller loss:
        \State \quad $g_t = \sum_{\tau=0}^{H-1} c(x_\tau^{(t)}, u_\tau^{(t)}, \tilde{x}_\tau^{(t)})$
        \State Calculate controller loss gradient according to~\eqref{eq:policy_gradient_estimate}
        \State Estimate controller loss Hessian:
        \State \quad $\Lambda_t = \widehat{\nabla}_\phi g_t \widehat{\nabla}_\phi g_t^{\top} + \alpha \frac{\partial u_{0:H}}{\partial \phi} \frac{\partial u_{0:H}}{\partial \phi}^{\top} + \epsilon \mathbf{I}$
        \State Policy update:
        \State \quad $\phi_{t+1} \gets \phi_t - \eta_t \Lambda_t^{-1} \widehat{\nabla}_\phi g_t$
      \EndFor
  \end{algorithmic}
\end{algorithm}

The proposed algorithm initializes both the dynamics model $f_\theta$ and the control policy $\pi_\phi$ randomly.
In each episode $t$, we rollout the control policy for a fixed horizon $H$, and extend the buffer $\mathcal{D}$ by the state-action-next-state tuples. Similar to many existing model-based \ac{RL} algorithms, we approximate the dynamics model by minimizing the mean squared error for single-step prediction:
\begin{equation}
  \ell\left(\theta\right) = \mathbb{E}_{\left(x, u, x^+\right) \sim \mathcal{D}}\|f_\theta(x, u) - x^+\|^2. \label{eq:model_learning_loss}
\end{equation}
In practice, we sample a mini-batch $\left\{(x_i, u_i, x_{i+1})\right\}_i$ from the buffer $\mathcal{D}_{t+1}$ at episode $t$.

After updating the model, we compute an analytical policy gradient on the policy parameter $\phi$ using the closed-loop gradient.
Recall that the cost function for the trajectory rollout in episode $t$ is defined as:
\begin{gather}
  g\left(\phi^{(t)}; \tilde{\bm{x}}^{(t)}\right) = \sum_{k=0}^{H-1} c(x_{\tau}^{(t)}, u_{\tau}^{(t)}, \tilde{x}_{\tau}^{(t)}) \label{eq:policy_loss_func_episode}\\
  u_{\tau}^{(t)} = \pi_{\phi^{(t)}}(x_{\tau}^{(t)}, \tilde{x}_{\tau}^{(t)}), \  x_{{\tau}+1}^{(t)} = f(x_{\tau}^{(t)}, u_{\tau}^{(t)}). \notag
\end{gather}
While the true dynamics model $f$ is unknown, we use the learned dynamics model $f_\theta$ instead in the calculation.
We first compute block Jacobian matrices $A_t \in \mathbb{R}^{nH\times nH}$, $B_t \in \mathbb{R}^{nH\times mH}$, and $K_t \in \mathbb{R}^{mH\times nH}$, each composed of $H \times H$ blocks:
\begin{align*}
\left[A_t\right]_{ij} & =
  \begin{cases} \frac{\partial f_\theta (x_{j}^{(t)}, u_{j}^{(t)})}{\partial x_{j}^{(t)}}, & i=j+1, \\ \bm{0}, & \text{otherwise}, \end{cases}\\
\left[B_t\right]_{ij} & =
  \begin{cases} \frac{\partial f_\theta (x_{j}^{(t)}, u_{j}^{(t)})}{\partial u_{j}^{(t)}}, & i=j+1, \\ \bm{0}, & \text{otherwise}, \end{cases} \\
\left[K_t\right]_{ij} & =
  \begin{cases} \frac{\partial \pi_\phi (x_{j}^{(t)}, \tilde{x}_{j}^{(t)})}{\partial x_{j}^{(t)}}, & i=j, \\ \bm{0}, & \text{otherwise}, \end{cases}
\end{align*}
for $i, j \in \left\{0, \ldots, H-1\right\}$.
Denote $\bm{x}_t = \mathrm{vec} (x_0^{(t)}, \ldots, x_{H-1}^{(t)})$ and $\bm{u}_t = \mathrm{vec} (u_0^{(t)}, \ldots, u_{H-1}^{(t)})$ as the concatenated state and control trajectories of the rollout in episode $t$.
Then, we estimate the policy gradient as:
\begin{multline}
  \widehat{\nabla}_\phi g\left(\phi_t; \tilde{\bm{x}}_t\right) = \\
  \frac{\partial g\left(\phi_t; \tilde{\bm{x}}_t\right)}{\partial \bm{x}_{t}}
  \left(\bm{\mathrm{I}} - \left(A_t + B_t K_t \right)\right)^{-1}
  B_t \frac{\partial \bm{u}_{t}}{\partial \phi},
  \label{eq:policy_gradient_estimate}
\end{multline}
where $\widehat{\nabla}_\phi g\left(\phi_t; \tilde{\bm{x}}_t\right)$ captures the closed-loop sensitivity of the cost $g\left(\phi_t; \tilde{\bm{x}}_t\right)$ with respect to the policy parameters $\phi_t$, taking into account the feedback effects through the system dynamics and the control policy.
As $A_t + B_t K_t$ is only non-zero on the first upper diagonal blocks, the matrix inversion in~\eqref{eq:policy_gradient_estimate} essentially unrolls the closed-loop dynamics over time.

The calculation in~\eqref{eq:policy_gradient_estimate} is different from common implementations of back-propagation through time (BPTT) algorithm, which typically uses differentiable dynamics for forward simulation and computes the gradients using a backward pass through the computation graph~\citep{Zhang25LearningVisionbased}.
This is not possible in our case, as the real dynamics are unknown.
Yet the Jacobian of the learned dynamics model $f_\theta$ can be used to evaluate an approximate gradient along the actual rollout trajectory.
While the learned model may be inexact, i.e., $f_\theta (x_\tau, u_\tau) \neq x_{\tau+1}$, the input-output Jacobians of $f_\theta$ can still provide useful information about the local dynamics along the true trajectory.
With this technique, we avoid rolling out the learned model to collect trajectories, which would be affected by compounding errors, and gain significant computational efficiency as we can use first-order optimization methods on real trajectories.

The policy optimization in Alg.~\ref{alg:online_mbrl} uses a pre-conditioned online gradient descent algorithm:
\begin{equation}
  \phi_{t+1} = \phi_t - \eta_t \Lambda_t^{-1} \widehat{\nabla}_\phi g\left(\phi_t; \tilde{\bm{x}}_t\right), 
  \label{eq:policy_update_step}
\end{equation}
where $\eta_t$ is the learning rate, and the pre-conditioner $\Lambda_t$ is defined as:
\begin{equation}
   \widehat{\nabla}_\phi g\left(\phi_t; \tilde{\bm{x}}_t\right) \widehat{\nabla}_\phi g\left(\phi_t; \tilde{\bm{x}}_t\right)^{\top} + \alpha \frac{\partial u_{0:H}}{\partial \phi} \frac{\partial u_{0:H}}{\partial \phi}^{\top} + \epsilon \mathbf{I}.
  \label{eq:preconditioner}
\end{equation}


The policy optimization step \eqref{eq:policy_update_step} solves an optimization problem with linearized objective and $\Lambda_t$-norm regularization.
As $\pi_\phi$ and $f_\theta$ are both highly nonlinear functions, the estimated gradient $\widehat{\nabla}_\phi g\left(\phi_t; \tilde{\bm{x}}_t\right)$ is likely to be inaccurate once the policy changes significantly, thus the regularization is necessary to stabilize the learning algorithm.
The three terms in~\eqref{eq:preconditioner}, controlled by the hyperparameters $\alpha$ and $\epsilon$, restrict policy parameter updates that cause significant changes in the objective function, control actions, and policy parameters, respectively.
A similar regularization has been used in~\citep{Ma25StochasticOnline}, where it was interpreted as a trust region approach to prevent the policy from changing too much in each episode.

\subsection{Theoretical Analysis}
We analyze the proposed algorithm from an online \revision{optimization} perspective and justify that the control policy converges under necessary assumptions and appropriate selection of the hyperparameters.
Algorithm~\ref{alg:online_mbrl} can be viewed as simultaneously solving two interleaved online \revision{optimization} problems: the online model learning and the online policy learning.
Specifically, the online policy learning relies on the learned dynamics model to compute an approximate policy gradient, while the online model learning is affected by policy updates because updated policies induce shifting data distributions in the buffer $\mathcal{D}$.
While formally analyzing the joint dynamics of the two learning problems is difficult, we separately analyze the two problems under necessary assumptions.
Each problem is analyzed as a special case of online \revision{optimization}, and in the end, we discuss how joint convergence can be achieved for Algorithm~\ref{alg:online_mbrl}.

\subsubsection{Online Policy Learning}
The online policy learning can be viewed as an online optimization problem with an inexact loss gradient.
We let $\phi^{\star} = \arg\min_{\phi\in\Phi} \mathbb{E}_{\tilde{\bm{x}}\sim\tilde{D}} [g(\phi;\tilde{\bm{x}})]$ and define the expected regret for the policy learning as:
\begin{equation}
  \mathbb{E}\left[\mathrm{R}_T\right] \doteq \mathbb{E}\left[\sum_{t=1}^T g\left(\phi_t; \tilde{\bm{x}}_t\right) - g\left(\phi^{\star}; \tilde{\bm{x}}_t\right)\right]. \label{eq:policy_regret_sum}
\end{equation}

Furthermore, we assume that the cost function $g$ is $L_g$-Lipschitz continuous and $\beta_g$-smooth for any $\tilde{\bm{x}} \in \widetilde{\mathcal{X}}$ and $\phi,~\phi^{\prime} \in \Phi$.
Then we have
\begin{equation*}
g(\phi^{\prime}; \tilde{\bm{x}}) \leq g(\phi; \tilde{\bm{x}}) + \langle \nabla_{\phi} g(\phi; \tilde{\bm{x}}), \phi^{\prime}  - \phi \rangle + \frac{\beta_g}{2} \|\phi^{\prime}  - \phi\|^2.
\end{equation*}
Plugging in the policy update rule~\eqref{eq:policy_update_step}, we get,
\begin{multline*}
g(\phi_{t+1}; \tilde{\bm{x}}) \leq g(\phi_t; \tilde{\bm{x}}) - \eta_t \langle \nabla_{\phi} g(\phi_t; \tilde{\bm{x}}), \Lambda_t^{-1} \widehat{\nabla}_\phi g(\phi_t; \tilde{\bm{x}}) \rangle \\
+ \frac{\beta_g \eta_t^2}{2} \|\Lambda_t^{-1} \widehat{\nabla}_\phi g(\phi_t; \tilde{\bm{x}})\|^2.
\end{multline*}

We denote the gap between the true gradient and the approximate gradient as $\delta_t \doteq \widehat{\nabla}_\phi g(\phi_t; \tilde{\bm{x}}) - \nabla_{\phi} g(\phi_t; \tilde{\bm{x}})$, and we can rewrite the above equation as
\begin{multline}
g(\phi_{t+1}; \tilde{\bm{x}}) \leq  g(\phi_t; \tilde{\bm{x}}) \\
- \eta_t \langle \nabla_{\phi} g(\phi_t; \tilde{\bm{x}}), \Lambda_t^{-1} \nabla_\phi g(\phi_t; \tilde{\bm{x}}) \rangle \\
 - \eta_t \langle \nabla_{\phi} g(\phi_t; \tilde{\bm{x}}), \Lambda_t^{-1} \delta_t \rangle \\
 + \frac{\beta_g \eta_t^2}{2} \|\Lambda_t^{-1} \widehat{\nabla}_\phi g(\phi_t; \tilde{\bm{x}})\|^2.
\label{eq:policy_perstep_loss_decomposition}
\end{multline}

By the definition~\eqref{eq:preconditioner}, $\Lambda_t$ is a positive definite matrix due to the sum of two outer products and a scaled identity matrix. Thus, the eigenvalues of $\Lambda_t^{-1}$ are well-defined and bounded as long as $\widehat{\nabla}_\phi g(\phi_t; \tilde{\bm{x}})$ and $\nicefrac{\partial u_{0:H}}{\partial \phi}$ have bounded norms.
We assume this property holds, and the eigenvalues of $\Lambda_t$ are bounded by
\begin{equation*}
  0 < m \leq \lambda_{\min}(\Lambda_t) \leq \lambda_{\max}(\Lambda_t) \leq M,
\end{equation*}
for all $t \geq 0$. Therefore, we can bound the first term in~\eqref{eq:policy_perstep_loss_decomposition} as follows
\begin{equation*}
\langle \nabla_{\phi} g(\phi_t; \tilde{\bm{x}}), \Lambda_t^{-1} \nabla_\phi g(\phi_t; \tilde{\bm{x}}) \rangle \geq \frac{1}{M} \|\nabla_\phi g(\phi_t; \tilde{\bm{x}})\|^2.
\end{equation*}
The second term in~\eqref{eq:policy_perstep_loss_decomposition} can be bounded using the Cauchy-Schwarz inequality:
\begin{multline*}
\left|\langle \nabla_{\phi} g(\phi_t; \tilde{\bm{x}}), \Lambda_t^{-1} \delta_t \rangle \right|\\
\leq \|\nabla_{\phi} g(\phi_t; \tilde{\bm{x}})\| \cdot \|\Lambda_t^{-1} \delta_t\| \leq  \frac{L_g}{m} \|\delta_t\|.
\end{multline*}
The last term in~\eqref{eq:policy_perstep_loss_decomposition} can be bounded by the Lipschitz property of $g$:
\begin{equation*}
\|\Lambda_t^{-1} \widehat{\nabla}_\phi g(\phi_t; \tilde{\bm{x}})\|^2 \leq \frac{1}{m^2} \|\widehat{\nabla}_\phi g(\phi_t; \tilde{\bm{x}})\|^2  \leq \frac{L_g^2}{m^2}.
\end{equation*}
Putting the bounds together, we have
\begin{multline}
 g(\phi_{t+1}; \tilde{\bm{x}}) \leq  g(\phi_t; \tilde{\bm{x}}) - \frac{\eta_t}{M} \|\nabla_\phi g(\phi_t; \tilde{\bm{x}}) \|^2 \\
+ \frac{\eta_t L_g}{m} \|\delta_t\|  + \frac{\beta_g \eta_t^2 L_g^2}{2m^2}. 
\label{eq:policy_loss_recursion_0}
\end{multline}

We note that~\eqref{eq:policy_loss_recursion_0} provides a bound for the loss change in one update of the policy parameter.
While the term $-\|\nabla_\phi g(\phi_t; \tilde{\bm{x}})\|^2$ is always non-positive, the gradient estimation error $\delta_t$ needs to be controlled to ensure the convergence. We provide one method to derive the requirement on $\delta_t$ by assuming a \ac{PL} condition on the loss function $g$ for any $\tilde{\bm{x}} \in \widetilde{\mathcal{D}}$ and $\phi \in \Phi$
\begin{equation*}
  \|\nabla_\phi g(\phi; \tilde{\bm{x}})\|^2 \geq 2\mu (g(\phi; \tilde{\bm{x}}) - g(\phi^{\star}; \tilde{\bm{x}})), \quad \mu > 0.
\end{equation*}
The \ac{PL} assumption relates the gradient norm to the function value gap compared to the optimal parameters.
In our case, $g(\phi_t; \tilde{\bm{x}}) - g(\phi^{\star}; \tilde{\bm{x}})$ is exactly the regret in episode $t$.
Therefore we could rewrite~\eqref{eq:policy_loss_recursion_0} in the following form
\begin{multline}
 g(\phi_{t+1}; \tilde{\bm{x}}) - g(\phi^{\star}; \tilde{\bm{x}}) \\
\leq  \left(1 - \frac{2\mu\eta_t}{M}\right) \left(g(\phi_t; \tilde{\bm{x}}) - g(\phi^{\star}; \tilde{\bm{x}})\right) \\
+ \frac{\eta_t L_g}{m} \|\delta_t\| + \frac{\beta_g \eta_t^2 L_g^2}{2m^2}.
\label{eq:policy_loss_recursion_1}
\end{multline}

We let $G\left(\phi\right) \doteq \mathbb{E}\left[g\left(\phi; \tilde{\bm{x}}\right)\right]$ and take the expectation on both sides to obtain a recursive inequality for the expected regret:
\begin{multline}
 G(\phi_{t+1}) - G(\phi^*)  \\
\leq \left(1 - \frac{2\mu\eta_t}{M} \right) \left( G(\phi_{t}) - G(\phi^*) \right) \\
+ \mathbb{E} \left[\frac{\eta_{t} L_g}{m} \|\delta_{t}\| \right] + \frac{\beta_g \eta_{t}^2 L_g^2}{2m^2}.
\label{eq:policy_loss_recursion_2}
\end{multline}
This recursion suggests that the expected regret in each episode has a decreasing term driven by the online gradient descent as long as $\eta_t < \nicefrac{M}{2\mu}$.
However, an error term appears due to the inexact gradient, in addition to the term from higher-order residuals.
One can see that a necessary condition for the learned policy to achieve optimal performance is to have the gradient estimation error $\delta_t$ converge to $0$.

To derive a sufficient condition on $\delta_t$, we unroll the recursion in~\eqref{eq:policy_loss_recursion_2}.
Let $\eta_t = \eta$ for all $t$ with $1-\frac{2\mu\eta}{M} \in (0, 1)$, and assuming $G(\phi_0) - G(\phi^*) \leq D$ for some constant $D$, the previous expression can be simplified to
\begin{multline*}
 G(\phi_{t}) - G(\phi^*)  
\leq \left(1-\frac{2\mu\eta}{M}\right)^{t} D \\
 + \frac{\eta L_g}{m} \mathbb{E} \left[ \sum_{\tau=0}^{t-1} \left(1-\frac{2\mu\eta}{M}\right)^{t-\tau} \|\delta_{\tau}\| \right] \\
+ \frac{\beta_g \eta^2 L_g^2}{2m^2} \sum_{\tau=0}^{t-1} \left(1-\frac{2\mu\eta}{M}\right)^{t-\tau}.
\end{multline*}

Next, we sum over $t=1, 2, \dots, T$, we have
\begin{multline*}
\mathbb{E} \left[ \mathrm{R}_T \right] = \mathbb{E} \left[ \sum_{t=1}^{T} g(\phi_t) -  \sum_{t=1}^{T} g(\phi^*) \right] \\
\leq  \underbrace{\sum_{t=1}^{T} \left(1-\frac{2\mu\eta}{M}\right)^{t} D}_{\left(\mathrm{A}\right)} \\
+ \underbrace{\frac{\eta L_g}{m} \sum_{t=1}^{T} \mathbb{E} \left[ \sum_{\tau=0}^{t-1} \left(1-\frac{2\mu\eta}{M}\right)^{t-\tau} \|\delta_{\tau}\| \right]}_{\left(\mathrm{B}\right)} \\
+ \underbrace{\frac{\beta_g \eta^2 L_g^2}{2m^2} \sum_{t=1}^{T} \sum_{\tau=0}^{t-1} \left(1-\frac{2\mu\eta}{M}\right)^{t-\tau}}_{\left(\mathrm{C}\right)}
\end{multline*}

From the sum of a geometric series, for large enough $T$, we have
\begin{equation*}
\left(\mathrm{A}\right)  \leq D \cdot \frac{M}{2\mu\eta} = \mathcal{O}(\eta^{-1}).
\end{equation*}
We rearrange the sums in $(\mathrm{B})$, 
{
\allowdisplaybreaks
\begin{multline*}
\left(\mathrm{B}\right) = \frac{\eta L_g}{m} \mathbb{E} \left[ \sum_{t=1}^{T} \sum_{\tau=0}^{t-1} \left(1-\frac{2\mu\eta}{M}\right)^{t-\tau} \|\delta_{\tau}\| \right] \\
\leq \frac{\eta L_g}{m} \mathbb{E} \left[ \sum_{\tau=0}^{T-1}  \sum_{t=\tau+1}^{T} \left(1-\frac{2\mu\eta}{M}\right)^{t-\tau} \|\delta_{\tau}\| \right] \\
\leq \frac{\eta L_g}{m} \frac{M}{2\mu\eta} \left(1 - \left(1-\frac{2\mu\eta}{M}\right)^{T}\right) \mathbb{E} \left[ \sum_{\tau=0}^{T-1} \|\delta_{\tau}\| \right].
\end{multline*}
}
By Cauchy-Schwarz and Jensen's inequality, the last term
\begin{equation*}
\mathbb{E} \left[ \sum_{\tau=0}^{T-1} \|\delta_{\tau}\| \right] \leq \sqrt{T \cdot \sum_{\tau=0}^{T-1} \|\delta_{\tau}\|^2} \leq \sqrt{ T \cdot \mathbb{E} \left[ \sum_{\tau=0}^{T-1} \|\delta_{\tau}\|^2 \right] },
\end{equation*}
and the bound on $(\mathrm{B})$ simplifies to
\begin{align*}
\left(\mathrm{B}\right) & \leq C \cdot \sqrt{T} \sqrt{\mathbb{E}\left[ \sum_{\tau=0}^{T-1} \|\delta_{\tau}\|^2 \right]},\ C = \frac{\eta L_g M}{2\mu m}.
\end{align*}
Finally, for the term (C), we have
\begin{equation*}
\left(\mathrm{C}\right) \leq \frac{\beta_g \eta^2 L_g^2}{2m^2} \frac{MT}{2\mu\eta} = \mathcal{O} \left( \eta T \right).
\end{equation*}

Combining all the terms, we have
\begin{equation}
\mathbb{E} \left[ \mathrm{R}_T \right] \leq \mathcal{O} \left( \eta^{-1} \right) + \mathcal{O} \left(\eta T \right) + C\cdot\sqrt{T\mathbb{E}\left[ \sum_{\tau=0}^{T-1} \|\delta_{\tau}\|^2 \right]}. 
\label{eq:policy_regret_final_bound_model_residual}
\end{equation}
The final regret bound~\eqref{eq:policy_regret_final_bound_model_residual} consists of the standard terms in online optimization, and an additional term that depends on the cumulative gradient prediction error $\mathbb{E}\left[\sum_{t=0}^{T-1} \|\delta_t\|^2\right]$.
The first two terms can be balanced to $\sqrt{T}$ by setting $\eta = \mathcal{O}(T^{-1/2})$, while the last term depends on the quality of the learned dynamics model.
Some special cases of~\eqref{eq:policy_regret_final_bound_model_residual} have been studied in the literature.
If the exact model is available (i.e., $\delta_t = 0$), the regret bound reduces to $\mathcal{O}(\sqrt{T})$, recovering the standard online convex optimization result.
If one applies the proposed algorithm with a prebuilt dynamics model that guarantees gradient prediction error bounded by a constant $\|\delta_t\| \leq \delta$, then the last term is $\mathcal{O}(T)$, leading to a linear regret.
If additional assumptions on the direction of predicted gradient are made, one can still achieve sublinear regret by controlling the cross-product term in~\eqref{eq:policy_perstep_loss_decomposition}.
This case has been thoroughly studied in~\citep{Ma25StochasticOnline}.

More generally,~\eqref{eq:policy_regret_final_bound_model_residual} suggests a sufficient condition for policy learning to achieve sublinear regret is to have $\mathbb{E}\left[\sum_{t=0}^{T-1} \|\delta_t\|^2\right] = o(T)$.
In Algorithm~\ref{alg:online_mbrl}, this requires the online model learning to converge sufficiently fast.

\subsubsection{Online Model Learning}
The previous derivation~\eqref{eq:policy_regret_final_bound_model_residual} shows that the regret of the policy learning depends on the expected cumulative gradient prediction error.
Before proceeding to analyze the online model learning problem itself, we first clarify the relationship between the model prediction error and the gradient prediction error.

As we do not have access to the true dynamics model or its gradient, the model learning part in Algorithm~\ref{alg:online_mbrl} is performed to minimize the one-step prediction error defined in~\eqref{eq:model_learning_loss}.
We justify that a model learned to accurately perform one-step prediction also has a small Jacobian prediction error under appropriate assumptions.
Many physical dynamical systems, in particular those considered in robotics, show continuity and smoothness, i.e., small changes in the state and control inputs lead to small changes in the next state.
Therefore, if a learned dynamics model accurately predicts the one-step transitions on a certain distribution of state-action pairs, it is likely that the learned model's Jacobian is also close to the true dynamics' Jacobian.
This intuition requires the distribution of the data used for model learning to be sufficiently rich, e.g., it must have sufficient probability density in the neighborhood of the trajectory for which we compute the policy gradient.
A formal analysis of this property from the perspective of measure theory can be found in~\citep{Lin24OnlinePolicy}.
\revision{
For robotic applications, the training process often uses randomly sampled commands.
We found it sufficient to obtain a rich data distribution when combined with randomly initialized control policies.
This supports the use of the model Jacobian in policy gradient estimation, although the model is only trained to minimize prediction error.
}

The Jacobian of the learned model is used to compute the policy gradient using~\eqref{eq:policy_gradient_estimate}.
Therefore, the policy gradient prediction error $\delta_t$ depends not only on the model Jacobian error, but also on the sensitivity of the closed-loop system formed by the dynamics and the control policy.
To avoid the error getting amplified by the $\left(\bm{\mathrm{I}} - (A_t + B_t K_t)\right)^{-1}$ term in~\eqref{eq:policy_gradient_estimate}, the closed-loop system must be stable.
This property can be assumed to hold in the initial phase for randomly initialized control policies and dissipative dynamics.
A discounting factor on the $A_t + B_t K_t$ term could also be used to further ensure computational stability.

The online model learning in Alg.~\ref{alg:online_mbrl} is performed similarly to standard online stochastic gradient descent.
However, unlike common online learning settings where the data is i.i.d. sampled from a constant distribution, the data distribution in our case is affected by the current control policy.
Specifically, the online model learning is always performed on the buffer $\mathcal{D}_{t}$ containing data collected from episode $1$ to $t$, and only used to compute the policy gradient on the current episode trajectory.
To analyze the model learning convergence with this distribution shift, we denote the distribution of the data collected in episode $t$ as $\nu_t$, and the distribution of the data in the buffer $\mathcal{D}_t$.
The distribution $\nu_t$ can be slowly changing as the policy is updated in each episode.
We use $\Delta_t \doteq \|\nu_t - \nu_{t-1}\|_{\mathrm{TV}}$ to denote the \ac{TV} distance between $\nu_t$ and $\nu_{t-1}$.
We impose the assumptions that:
\begin{itemize}
  \item $f$, $f_\theta$, are Lipschitz continuous and smooth functions;
  \item $\mathcal{X}$ and $\mathcal{U}$ are bounded compact sets;
  \item the loss function in~\eqref{eq:model_learning_loss} satisfies a \ac{PL} condition.
\end{itemize}

The expected model learning regret is defined as
\begin{equation}
  \mathbb{E}\left[ \bar{\mathrm{R}}_T \right] = \sum_{t=1}^T \mathbb{E}_{z \sim \nu_t} \ell(\theta_t; z) - \sum_{t=1}^T \mathbb{E}_{z\sim \nu_t} \ell(\theta_t^{\diamond}; z), 
  \label{eq:model_learning_regret}
\end{equation}
where $\theta_t^{\diamond} = \arg\min_{\theta} \mathbb{E}_{z\sim \nu_t} \ell(\theta; z)$.
One can show that~\eqref{eq:model_learning_regret} is bounded by the learning rate $\alpha$ and $\Delta_t,\ t=1,2,\ldots,T$ as
\begin{multline}
  \mathbb{E}\left[ \bar{\mathrm{R}}_T\right] \leq C_0 \sum_{t=1}^T \frac{1}{t} \sum_{i=2}^t (i-1)\Delta_i \\
  + \frac{C_1}{\alpha} 4K\sum_{t=1}^T \Delta_t + C_2 \alpha T + \frac{C_3}{\alpha},
  \label{eq:model_learning_regret_bound_with_delta}
\end{multline}
where $C_0, C_1, C_2, C_3$ are constants.
Similar to the analysis for policy learning in~\eqref{eq:policy_regret_sum}-\eqref{eq:policy_regret_final_bound_model_residual}, one can decompose the online gradient descent steps and isolate the disturbance caused by the shifting data distribution.
The detailed derivation is left to the Appendix.


The bound~\eqref{eq:model_learning_regret_bound_with_delta} suggests the influence of the distribution shift $\Delta_t$ on the online model learning process is two-fold.
The first term, originally from $(\mathrm{A})$ and $(\mathrm{C})$, measures the performance gap for training with the buffered data but only performing predictions on the current episode data.
The second term, originally from $(\mathrm{B})$, is due to the shifting distribution of the training data.
As the training data distribution is the uniform mixture of all past episode data distributions, the distribution shift $\Delta_t$ is discounted by a factor of $\nicefrac{1}{t}$, suggesting a single episode of data having a different distribution from the rest has a diminishing effect on the model learning process as more data is collected.

One can also see that when the data distribution across episodes is stationary (i.e., $\Delta_t = 0$), the regret bound reduces to $\mathcal{O}(\sqrt{T})$ by taking the optimal $\alpha \propto \nicefrac{1}{\sqrt{T}}$, which is consistent with the standard online gradient descent result for convex problems.
In case the distribution shift is bounded by a constant, i.e., $\Delta_t \leq D$, the bound becomes $\mathcal{O}(T) + \mathcal{O}(\nicefrac{T}{\alpha}) + \mathcal{O}(\alpha T) + \mathcal{O}(\nicefrac{1}{\alpha})$, and sublinear regret cannot be achieved.

\subsubsection{Summary}
The analysis above has separately considered the online \revision{optimization} problems for model learning and policy learning.
In practice, the two problems are coupled through the gradient prediction error $\delta_t$ and the data distribution shift $\Delta_t$, which creates difficulties for formal analysis.

In Algorithm~\ref{alg:online_mbrl}, a strong regularization on the policy update is used to ensure small changes between two consecutive policies.
This reduces the distribution shift $\Delta_t$ caused by the changing policy, and the model learning can view the data distribution as approximately stationary after an initial burn-in period to enrich the data buffer.
After this, the model learning problem becomes near-stationary and achieves fast convergence.
Literatures~\citep{Bubeck15ConvexOptimization} have shown that for strongly convex or \ac{PL}-satisfying problems, the convergence rate of $\nicefrac{1}{t}$ could be achieved with appropriate learning rate selection.
Practically, we could also use adaptive gradient methods such as Adam~\citep{Kingma15AdamMethod} to accelerate the convergence.
Assuming this faster convergence rate is achieved for $\delta_t$, the remainder term in~\eqref{eq:policy_regret_final_bound_model_residual} can be absorbed by the $\sqrt{T}$ term.
This lead to an effect of time-scale separation between the model learning and policy learning processes.
A similar technique is also used in adaptive control~\citep{Liberzon16NonlinearAdaptive}.

\subsubsection{Remarks}
We analyze the algorithm by studying two isolated online learning problems: the online policy optimization with learned model, and the online model learning with shifting data distribution.

Most assumptions on bounded state and action spaces, Lipschitz continuity, and smoothness of the dynamics function are common to many robotic systems.
\revision{
Arguably, for nonlinear dynamical systems, the convexity of the loss function with respect to the model parameters cannot be guaranteed nor verified.
However, the PL condition in the above analysis is necessary to extend the convergence of local regret to sublinear guarantees for the standard regret $R_T$.
Relaxed version of the analysis, similar to~\citep{Lin24OnlinePolicy, Ma25StochasticOnline}, could still characterize the convergence to first-order stationary points by showing sublinear local regret $R_T^{\mathrm{local}}$.
Although in this case, a sublinear local regret cannot be directly linked to absolute metrics of policy performance.
}


The presented theoretical analysis is not limited to the proposed algorithm.
The model learning convergence analysis seamlessly extends to other model-based RL algorithms, as many of them have the same setting of data accumulation in a buffer and \ac{SGD} updates per episode.
The main difference is that the policy difference between two consecutive episodes is not necessarily small, which is more likely to cause instabilities.
In addition, the relationship between the model prediction regret bound and control policy regret bound in those algorithms needs to be analyzed separately.

The analysis for the policy training problem is closely related to the use of differentiable models to optimize the control policy.
As a widely used technique in a number of algorithms from iterative learning control~\citep{Pierallini25FishingData, Ma22LearningbasedIterative} to RL in differentiable simulation~\citep{Song25LearningQuadruped, Zhang25LearningVisionbased}, it has been shown that even inaccurate gradient information from a surrogate model can be used to effectively optimize control policies.
Previous analyses have shown that the convergence of approximate gradient-based algorithms is only guaranteed if the model always predicts the correct direction of the gradient~\citep{Ma25StochasticOnline}.
While models satisfying this guarantee may be hard to obtain for robotic systems with multiphysical dynamics, our analysis shows it is possible to mitigate this requirement with online model learning.
In particular, from the analysis of policy learning, we notice that for any model learning process with square-summable gradient prediction error, the policy optimization process can achieve a regret bound related to the model error.
This suggests that the sublinear regret bound for control policy learning can be achieved by combining the proposed online policy learning with any model learning process that achieves sublinear regret bounds.

Online optimization of multiple models in the same pipeline, combined with changing objectives and data distributions, is also commonly seen in many model-free \ac{RL} algorithms, such as the Q-function and policy in DDPG~\citep{Lillicrap15ContinuousControl}, and the actor and critic in SAC or PPO~\citep{Haarnoja18SoftActorCritic, Schulman17ProximalPolicy}.
A similar analysis can be applied to these approaches, potentially helping to interpret the instabilities and justify the necessity for regularization.

\section{Experiment}
\label{sec:results}

We evaluate the proposed algorithm on two real-world robotic platforms: a hydraulic robot excavator and a cable-driven soft robot arm.
Both systems present distinct challenges in control, as their actuation dynamics are highly nonlinear, difficult to model, and exhibit significant heterogeneity—each unit differs and requires separate modeling.
We applied the proposed online model-based \ac{RL} approach to all of them and successfully learned control policies that adapt to the system dynamics and achieve high control accuracy.

\subsection{HEAP}
\begin{figure}[bthp]
  \centering
  \includegraphics[width=\linewidth]{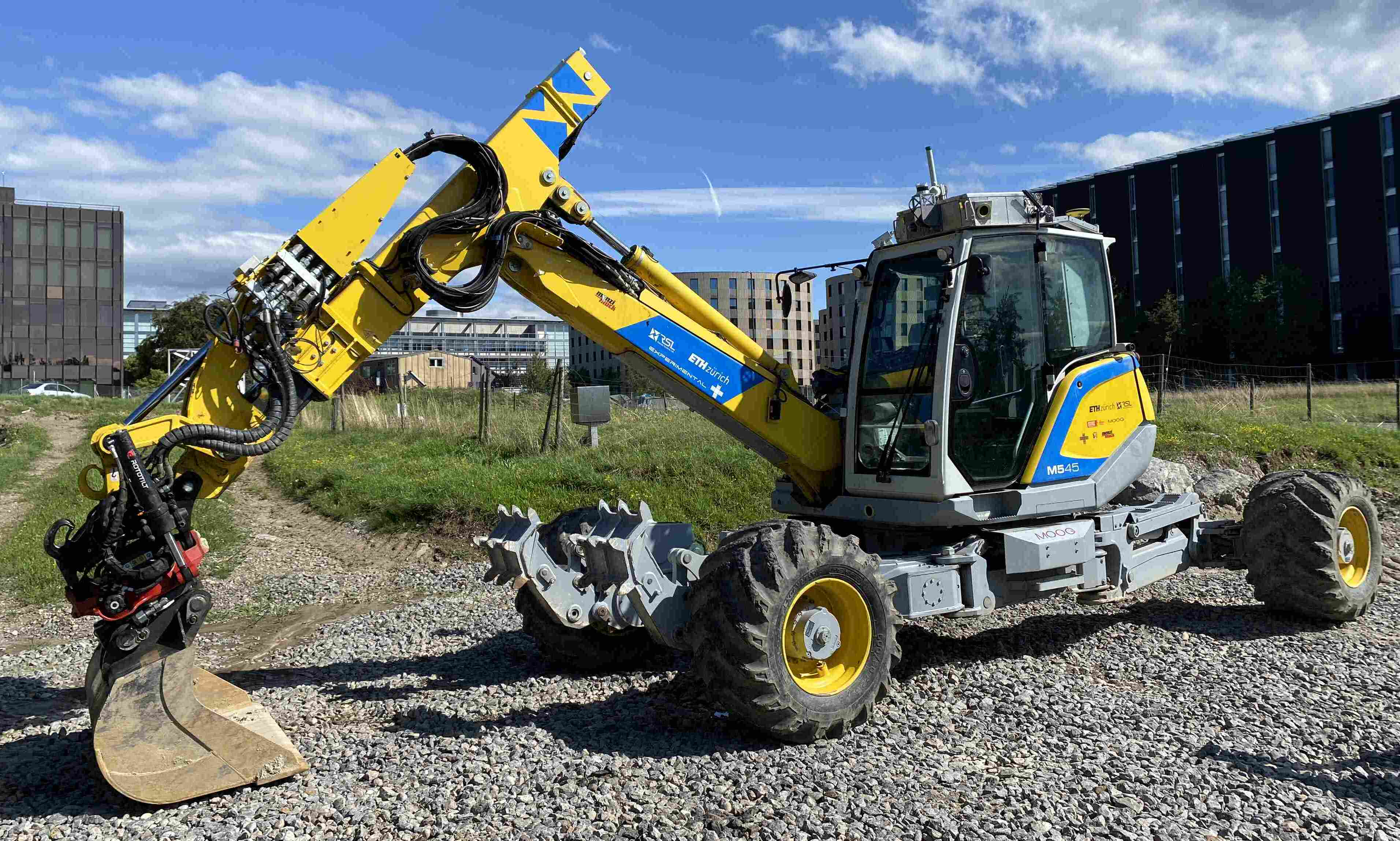}
  \caption{HEAP, the autonomous excavator used in the experiments.}
  \label{fig:picture_heap}
\end{figure}
HEAP is a 12.5-ton hydraulic excavator developed for autonomous construction purposes~\citep{Jud21HEAPAutonomous}.
It is equipped with computer-controlled proportional valves for each hydraulic actuator and high-precision proprioceptive sensors, enabling precise state estimation and automatic control.
Accurate control of the hydraulic-actuated arm of HEAP is crucial for tasks such as grading, digging, and loading.
While previous works to achieve this task always rely on some form of modeling, either through manual calibration and system identification~\citep{Jud21HEAPAutonomous} or data-driven modeling~\citep{Egli22GeneralApproach}, we apply the proposed online model-based \ac{RL} algorithm to obtain a control policy directly in one step.

The goal of the task is to control the excavator arm to follow a desired trajectory, which is defined by a sequence of target positions in the Cartesian space associated with time stamps.
The control policy is trained to minimize the tracking error, defined as the Euclidean distance between the current position and the target position at this time step.
The target trajectories are generated as polynomial functions with continuous first and second derivatives, ensuring smooth transitions.
Each trajectory lasts for 5 seconds, and each episode of training consists of 10 such trajectories.
As the system exhibits significant delay, we provide as input to the control policy not only the current position and velocity of the excavator arm joints but also a lookahead window of the target trajectory, along with the previous current commands.
The control policy is trained to predict a current command for each valve, which is then applied to the system.
The model is trained to predict the velocity of the arm at the next time step, which is then used to compute the next position.
Both the model and the control policy are represented by neural networks with two hidden layers.
Although both the model and the control policy are trained with online SGD in the analysis, we observe that using the Adam optimizer~\citep{Kingma15AdamMethod} for model learning significantly improves the convergence speed.
Therefore, we use Adam for model learning and SGD for control policy learning in the experiments.

\subsubsection{Simulation}
We first evaluate the proposed algorithm in a simulation environment of HEAP.
The simulation environment is built with the model created by~\citet{Egli22GeneralApproach}.
Trained with approximately \SI{100}{\minute} of data, the simulation uses a neural network to model the complex dynamics of HEAP's arm.

\begin{figure}[t]
  \centering
  \includegraphics[width=\linewidth]{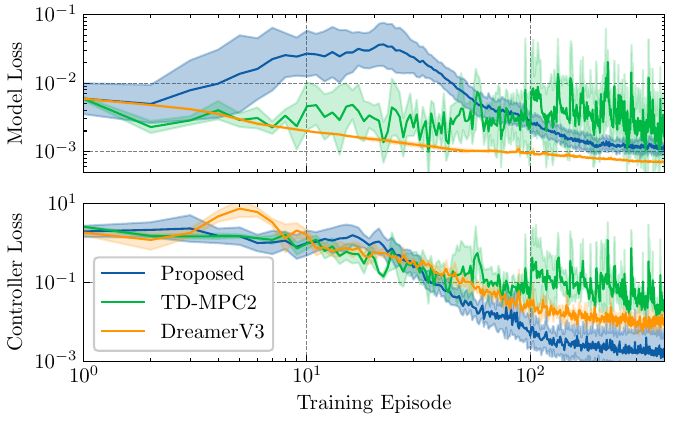}
  \caption{\revision{Performance of the proposed online model-based \ac{RL} and baseline algorithms in HEAP simulation. The plots show the average loss in model learning and control policy learning over multiple runs.} The shaded area represents the standard deviation. Both mean and standard deviation are computed on a log scale.
  \revision{The horizontal axis has been normalized to represent the same number of environment interactions across algorithms. The Model Loss curves represent the sum of all loss metrics for forward prediction quality and are normalized by the value after the first episode, to enable a qualitative comparison.}}
  \label{fig:heap_sim_loss}
\end{figure}

Deploying the proposed online model-based \ac{RL} algorithm in the simulation environment results in convergence of episodic loss in both model learning and control policy learning, as shown in Figure~\ref{fig:heap_sim_loss}.

\begin{figure}[t]
  \centering
  \begin{subfigure}[b]{0.24\textwidth}
    \includegraphics[width=\linewidth]{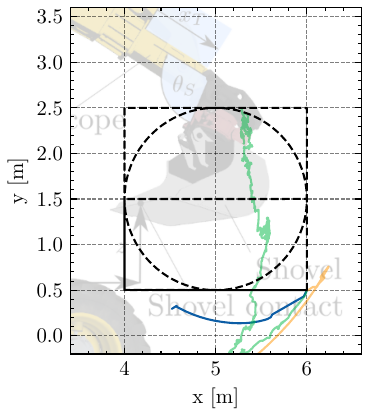}
    \caption{Episode 1}
    \label{fig:heap_realworld_traj_1}
  \end{subfigure}
  \hfill
  \begin{subfigure}[b]{0.24\textwidth}
    \includegraphics[width=\linewidth]{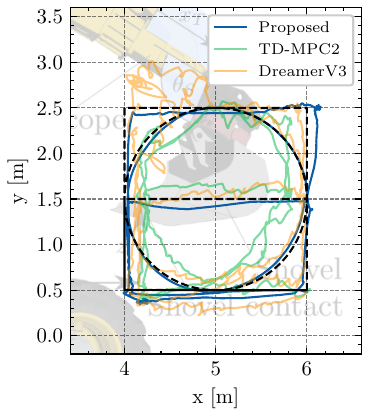}
    \caption{Episode 100}
    \label{fig:heap_realworld_traj_100}
  \end{subfigure}
  \begin{subfigure}[b]{0.24\textwidth}
    \includegraphics[width=\linewidth]{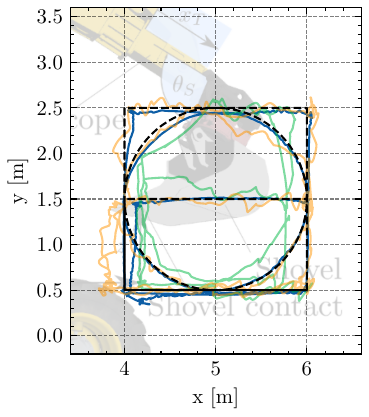}
    \caption{Episode 200}
    \label{fig:heap_realworld_traj_200}
  \end{subfigure}
  \hfill
  \begin{subfigure}[b]{0.24\textwidth}
    \includegraphics[width=\linewidth]{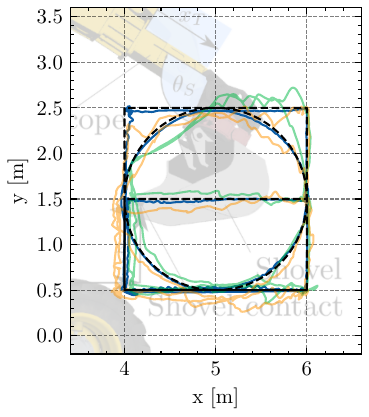}
    \caption{Episode 400}
    \label{fig:heap_realworld_traj_400}
  \end{subfigure}
  \caption{\revision{Evaluation of the control policy in simulation. The plots show the tracking error of the control policy with the proposed online MBRL algorithm and a baseline using TD-MPC2 on unseen reference trajectory that consists of straight lines and circles. The dashed line represents the reference.}}
  \label{fig:heap_sim_eval}
\end{figure}

The constant rate decreasing period in the log-scale loss plots indicates a power-law convergence behavior after the initial burn-in phase.
The burn-in phase is most likely caused by the limited data buffer size in the first few episodes, which results in a high variance in model learning and non-improving control policy performance.
After the first 10 episodes, a stable learning process is established, and the model and control policy losses decrease at a constant rate.
Measuring the slope of the policy loss in this segment, we estimate the convergence rate approximately $\ell_t \propto t^{-4}$.
After 200 episodes, the learning slows down significantly, and the loss reaches a plateau.
We suspect this is due to the complexity of the task limiting the achievable performance.

\revision{
  We perform the same experiment in simulation with two model-based RL baseline methods, TD-MPC2~\citep{Hansen24TDMPC2Scalable} and DreamerV3~\citep{Hafner23MasteringDiverse}.
  They represent two categories of model-based RL algorithms.
  TD-MPC2 trains a latent-space dynamics model along with reward and value predictors, and uses the learned model to perform receding horizon planning with a sampling-based method.
  DreamerV3 learns a similar dynamics model but uses the model to generate imagined rollouts, on which policy optimization is performed.
  Figure~\ref{fig:heap_sim_loss} shows the loss curves of the baseline algorithms on the same task.
  All algorithms were trained for the same amount of environment interactions, which suggests more than 100 times higher number of updates for the baseline algorithms as they learn per-step rather than per-episode.
  Nevertheless, both baseline algorithms perform worse than the proposed method.
  The controller performance of TD-MPC2 is significantly worse than that of the proposed algorithm, by at least an order of magnitude after 400 episodes of training.
  We further notice that TD-MPC2 has a much higher variance in the learning process, due to individual runs exhibiting random, sudden performance drops.
  We suspect that the performance drop is caused by the sampling-based planning method, which can show vastly different behavior between consecutive episodes, even following small changes in the learned model.
  DreamerV3, on the other hand, shows a stable decreasing loss in both model and policy performance, yet the policy performance after convergence is not as good as the proposed method.
  This suggests the policy trained on model-generated data still suffers from a domain gap, and for nonlinear control problems, the domain gap may not be improved even if the learned model is increasingly accurate.
  On the contrary, the proposed algorithm uses specifically designed online optimization updates and only real environment interaction data for policy optimization, which results in a more stable learning process and significantly better performance.
}

\revision{The controller performance of the proposed algorithm is evaluated on a set of fixed reference trajectories throughout the training and compared against control policies trained with baseline methods. The results are shown in Figure~\ref{fig:heap_sim_eval}.
The control policy achieves a mean tracking error of \SI{5.5}{\centi\meter} after only 200 episodes of training or equivalently \SI{2.78}{\hour} of training data experienced in simulation, and keeps improving with further training.
The baseline model-based RL algorithms also shows improving performance over time, although they do not reach the same performance.
}
On the other hand, model-free policy gradient methods such as \ac{PPO}, as reported by~\citet{Egli22GeneralApproach} \revision{takes more than \SI{10000}{\hour} of simulated experience to converge}.
\revision{This drastic reduction in training time and data usage, as well as superior controller performance, are key advantages of the proposed online model-based \ac{RL} algorithm, which allows us to directly deploy the proposed algorithm on the real robot for accurate controller training.}


\subsubsection{Real-World Deployment}
\begin{figure}[t]
  \centering
  \includegraphics[width=\linewidth]{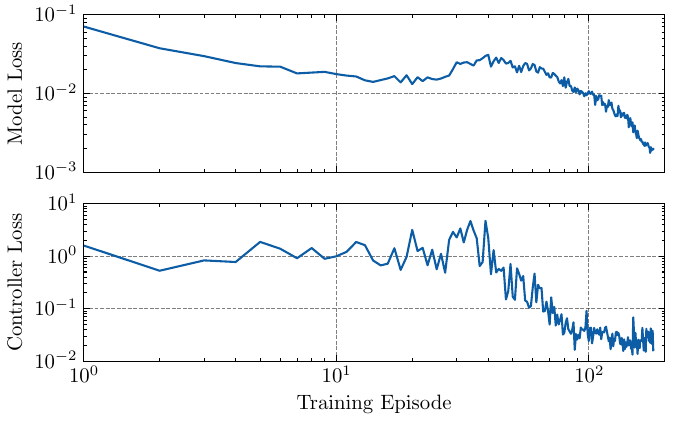}
  \caption{Real-world deployment results on the HEAP excavator. The plots show convergence, evaluation metrics, and adaptation to payload changes.}
  \label{fig:heap_realworld_loss}
\end{figure}

\begin{figure*}[t]
  \centering
  \begin{subfigure}[b]{0.24\textwidth}
    \includegraphics[width=\linewidth]{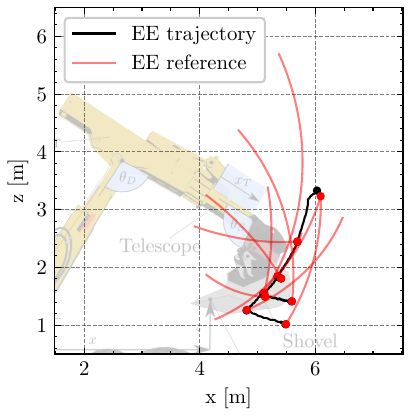}
    \caption{Episode 1}
    \label{fig:heap_realworld_traj_1}
  \end{subfigure}
  \hfill
  \begin{subfigure}[b]{0.24\textwidth}
    \includegraphics[width=\linewidth]{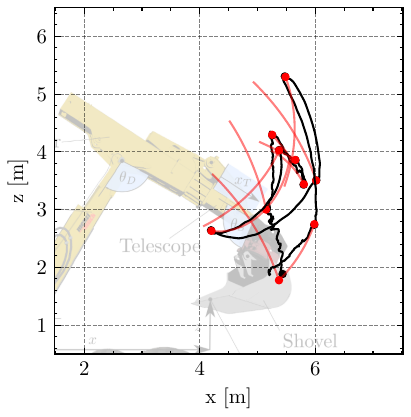}
    \caption{Episode 60}
    \label{fig:heap_realworld_traj_2}
  \end{subfigure}
  \hfill
  \begin{subfigure}[b]{0.24\textwidth}
    \includegraphics[width=\linewidth]{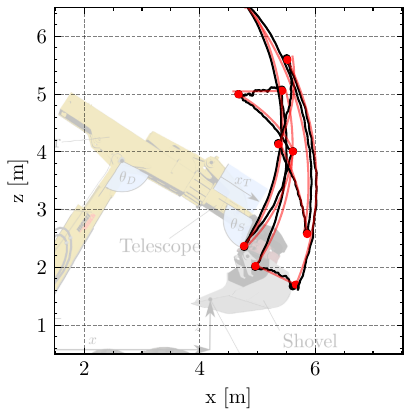}
    \caption{Episode 120}
    \label{fig:heap_realworld_traj_3}
  \end{subfigure}
  \hfill
  \begin{subfigure}[b]{0.24\textwidth}
    \includegraphics[width=\linewidth]{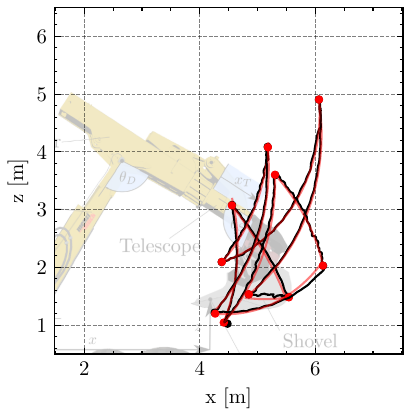}
    \caption{Episode 180}
    \label{fig:heap_realworld_traj_4}
  \end{subfigure}
  \caption{Real-world deployment results on the HEAP excavator at different training episodes. The plots show the tracking performance and adaptation over time.}
  \label{fig:heap_realworld_traj}
\end{figure*}

With the successful application of the proposed online model-based \ac{RL} algorithm in simulation, we deploy the algorithm on the real HEAP excavator.
We initialize the model and control policy randomly as done in the simulation experiment, and run the \revision{proposed} algorithm for 180 episodes.
This corresponds to \SI{2.5}{\hour} of real-world training data.
The total experiment time is approximately \qty{3}{\hour} with the algorithm running on a desktop workstation\footnote{The training time is recorded on a desktop workstation with an Intel i9-12900K CPU and an NVIDIA RTX 4090 GPU. The algorithm is fully implemented on the GPU. The workstation communicates through ROS with an onboard computer, which executes the control loop.}.

The losses of the training process are shown in Figure~\ref{fig:heap_realworld_loss}.
We also record the performance of the control policy over training episodes at different times and plot the results in Figure \ref{fig:heap_realworld_traj}.

The performance of the control policy reaches a mean tracking error of \SI{2.7}{\centi\meter} following random spline trajectories after 180 episodes.
A detailed comparison of the accuracy with previous approaches ~\citep{Egli22GeneralApproach, Nan24LearningAdaptive} is shown in Table~\ref{tab:heap_error_comparison}.
\revision{
Specifically, \citet{Egli22GeneralApproach} models the dynamics of HEAP by collecting data from the real system manually with additional excitation signals, and then trains control policies with \ac{RL} using the learned model as a simulator.
\citet{Nan24LearningAdaptive} uses \ac{RL} to train a control policy in a parameterized simulation environment along with an adaptation strategy that estimates the correct latent parameters, so that the policy can be adapted to the real system at test time.
}
Table~\ref{tab:heap_error_comparison} shows several metrics of all three methods, including the average velocity $v^{\mathrm{avg}}$, maximum velocity $v^{\mathrm{max}}$, average positional tracking error $|e_p|^{\mathrm{avg}}$, maximum positional tracking error $|e_p|^{\mathrm{max}}$, and the normalized performance indicator $\rho=|e_p|^{\mathrm{max}}/v^{\mathrm{max}}$ introduced in previous papers by~\cite{Mattila17SurveyControl}.

The online-learning-based approach achieves significantly higher velocity with similar or better normalized accuracy compared to previous approaches.
Table~\ref{tab:heap_error_comparison} lists the highest velocity performance reported by~\citet{Egli22GeneralApproach} and reproduced results of~\cite{Nan24LearningAdaptive} at higher speed.
Yet, the previous approaches cannot achieve similar accuracy even at half the achieved speed.
Specifically, $\rho$ reduced by $30\%$ when the maximum velocity more than doubled compared to~\cite{Egli22GeneralApproach}, which uses a data-driven model for the particular machine to train an \ac{RL} controller.
The reason is that the previous approach relies on the quality of the learned model, which heavily degrades at high speeds due to the limited training data.
While~\cite{Nan24LearningAdaptive} achieves higher velocity with its randomized pre-training and online adaptation approach, the accuracy is not comparable at high speed.
This is likely due to the fact that the randomized pre-training relies a simplified model of the hydraulic excavator, which has higher discrepancy with the real system because of the coupling between joints and load-dependent dynamics.
As the proposed algorithm directly learns from real-world data, these limitations are avoided, and the algorithm is able to achieve better accuracy at high speed.

\begin{table}[h]
  \centering
  \caption{Trajectory tracking performance and comparison with previous approach. The data of the baseline methods is taken from~\cite{Egli22GeneralApproach} and by reproducing the result in~\cite {Nan24LearningAdaptive} at a higher speed.}
  \defcitealias{Nan24LearningAdaptive}{N\&H24}
  \defcitealias{Egli22GeneralApproach}{E\&H22}
  \begin{tabular}{cccc}
      \toprule
      Metric & Proposed & \citepalias{Nan24LearningAdaptive} & \citepalias{Egli22GeneralApproach}\\
      \midrule
      $v^{\mathrm{avg}}$ [cm/s]              & 52.3  & 37.5  & 20    \\
      $v^{\mathrm{max}}$ [cm/s]              & 129.3 & 60    & 52.5  \\
      $|e_p|^{\mathrm{avg}}$ [cm]            & 4.5   & 8.56  & 2.8   \\
      $|e_p|^{\mathrm{max}}$ [cm]            & 12.4  & 16.2  & 6.7   \\
      $\rho$ [s]                             & 0.09  & 0.27  & 0.13  \\
      \bottomrule
  \end{tabular}
  \label{tab:heap_error_comparison}
\end{table}

\begin{figure}[h]
  \centering
  \includegraphics[width=\linewidth]{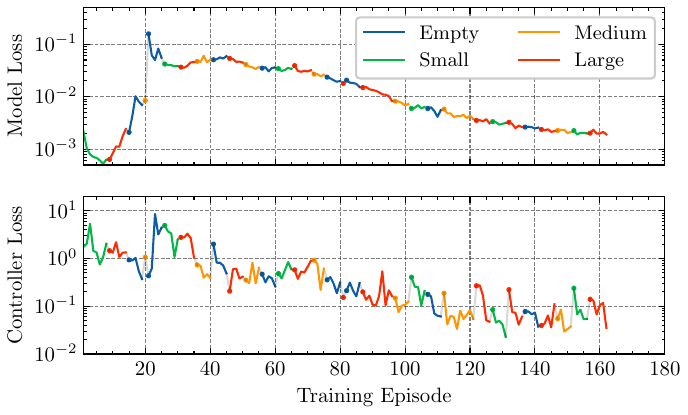}
  \caption{\revision{Experiment} on HEAP with changing payload conditions. The plots show loss evolution with different colors showing episodes collected under different payload conditions. The first episode after each payload change is highlighted with a dot.}
  \label{fig:heap_payload_change}
\end{figure}

\begin{figure}[h]
  \centering
  \begin{subfigure}[b]{0.32\linewidth}
    \includegraphics[width=\linewidth]{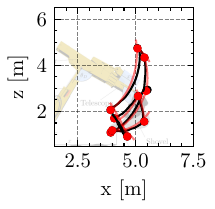}
    \caption{Episode 156}
    \label{fig:heap_realworld_rock_traj_1}
  \end{subfigure}
  \hfill
  \begin{subfigure}[b]{0.32\linewidth}
    \includegraphics[width=\linewidth]{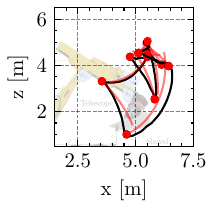}
    \caption{Episode 157}
    \label{fig:heap_realworld_rock_traj_2}
  \end{subfigure}
  \hfill
  \begin{subfigure}[b]{0.32\linewidth}
    \includegraphics[width=\linewidth]{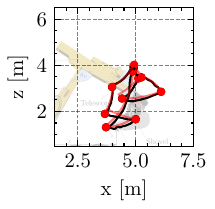}
    \caption{Episode 162}
    \label{fig:heap_realworld_rock_traj_3}
  \end{subfigure}
  \caption{An example of online \revision{adaptation} on HEAP with changing payload conditions. The plots show the tracking performance at (a) episode 156 with a small rock payload, (b) episode 157 right after switching to a large rock payload, and (c) episode 162 after a few episodes of adaptation.}
  \label{fig:heap_realworld_rock_traj}
\end{figure}

The robustness of the \revision{proposed} algorithm is also demonstrated by its ability to adapt to shifting dynamics.
We verify this by changing the payload of the excavator arm, taking the gripper end effector, and switching to boulders of different weights in the gripper.
During this process, the \revision{proposed} algorithm learns to perform the same trajectory tracking task without information about the payload change.
The \revision{proposed} algorithm is able to adapt to the disturbances and quickly learns to control the arm with the new payload, despite the fact that the underlying dynamics model has changed.
The loss plots with payload changes during training are shown in Figure~\ref{fig:heap_payload_change}.
It can be seen that, despite having significantly higher randomness in the data, the model and control policy losses still decrease over time.
After the first 100 episodes, we observe that the controller loss increases significantly when the payload changes, but quickly recovers to a lower level after a few episodes.
One example of the payload switch at episode 157 is shown in Figure~\ref{fig:heap_realworld_rock_traj}.
The plot shows the tracking performance at episode 156 when the algorithm has been trained with a small rock payload, episode 157 right after switching to a large rock payload, and episode 162 after a few episodes of adaptation.
After the payload switch, the tracking error immediately doubles, but quickly reduces back to a similar level as before after five episodes of training adaptation.
This shows that the \revision{proposed} algorithm is able to adapt to the new payload conditions, although the algorithm was not explicitly designed to handle such disturbances.

\subsection{Soft Robot Arm}
\begin{figure}[h!]
  \centering
  \includegraphics[width=0.75\linewidth]{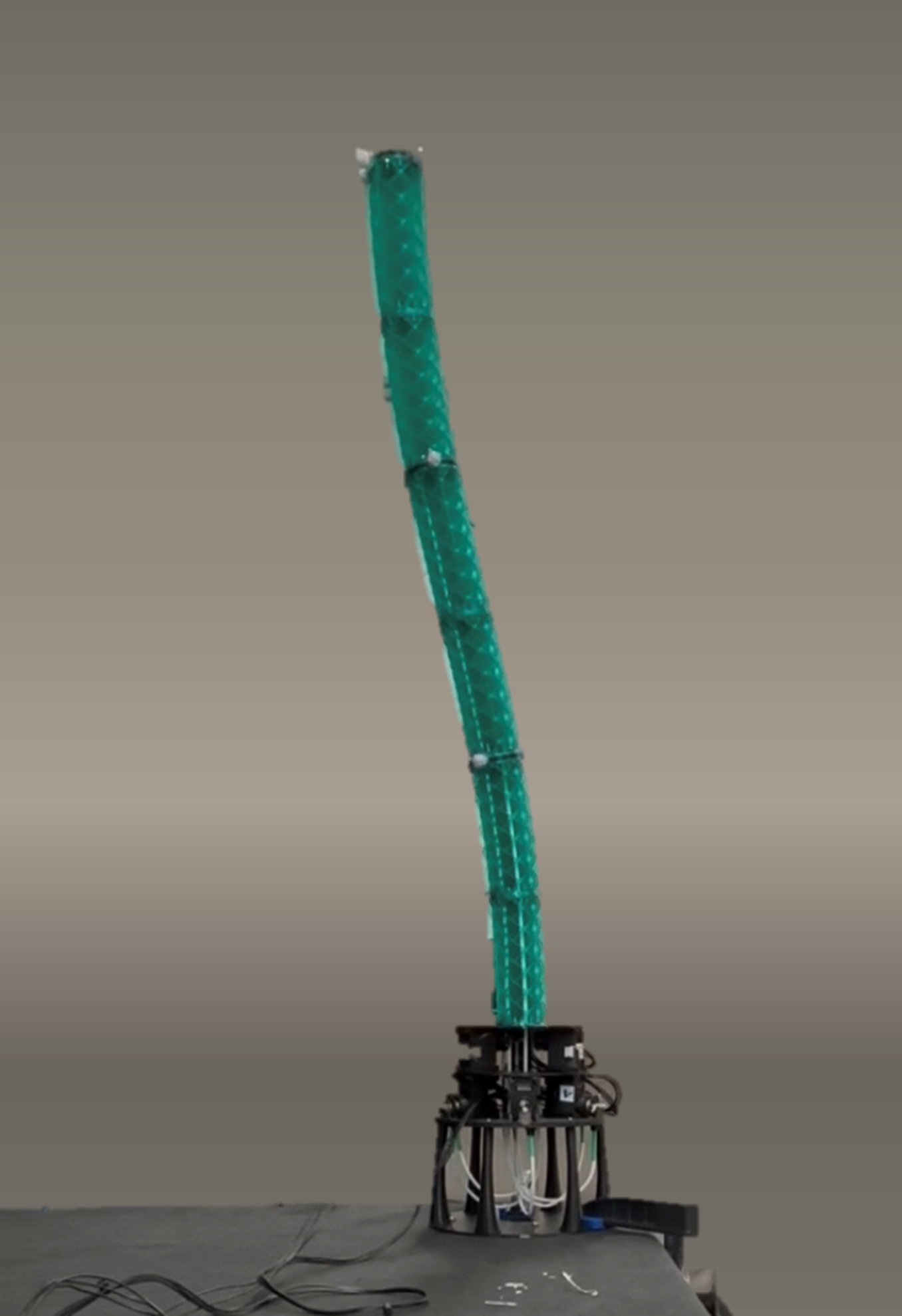}
  \caption{The cable-driven soft robot arm used in the experiments.}
  \label{fig:picture_soft_arm}
\end{figure}
The \revision{proposed} algorithm is further evaluated on a cable-driven soft robot arm shown in Figure~\ref{fig:picture_soft_arm}.
The hardware platform, designed by~\citet{Guan23TrimmedHelicoids}, consists of three helicoid segments, each driven by three cables connected to a spool driven by a motor in the base.
As the helicoid segment shall only be bent but not compressed, each segment can be controlled with two degrees of freedom.
An external motion capture system is used to track the poses of the ends of the helicoid segments.

We use the proposed algorithm to train a control policy that steers the tip of the robot arm to follow a desired trajectory in the $xy$-plane.
The learned model tries to predict the change in position of the end of the helicoid segments at the next time step, taking the state defined as the position of each helicoid segment's end.
During training, the reference trajectories are randomly generated as polynomial curves with continuous velocity profiles, and the control policy predicts the desired motor position command for two motors of each segment.
The third motor of each segment is controlled to maintain the centerline length constant.
Similar to the previous experiments, each episode consists of 10 trajectories, each lasting for 6 seconds.

\begin{figure}
  \centering
  \includegraphics[width=\linewidth]{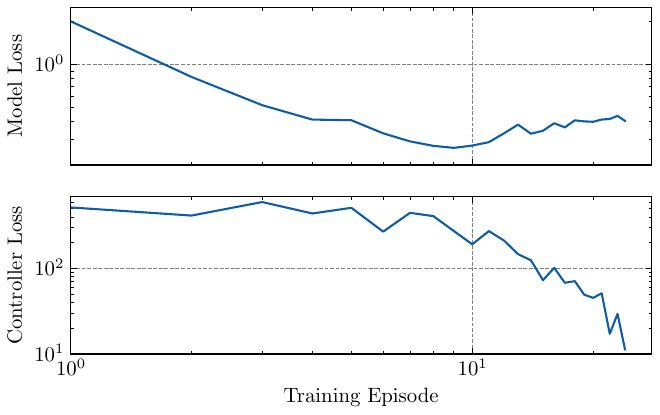}
  \caption{Convergence of the online model-based \ac{RL} algorithm on the soft robot arm.}
  \label{fig:soft_arm_loss}
\end{figure}

\begin{figure}
  \centering
  \begin{subfigure}[b]{\linewidth}
    \includegraphics[width=\linewidth]{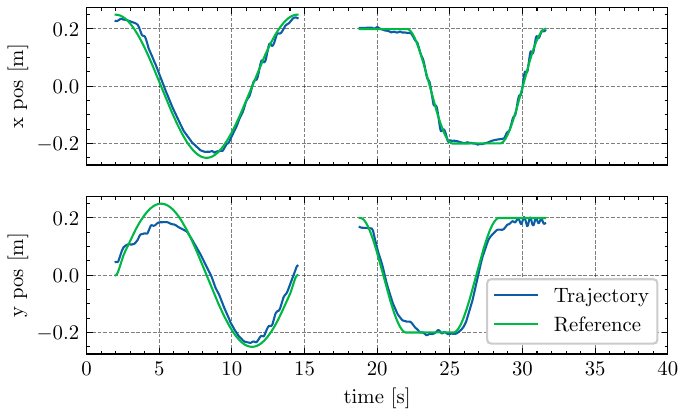}
    \caption{Reference and executed trajectory in $x$- and $y$-direction over time.}
    \label{fig:soft_arm_traj_xy_time}
  \end{subfigure}
  \hfill
  \begin{subfigure}[b]{\linewidth}
    \includegraphics[width=\linewidth]{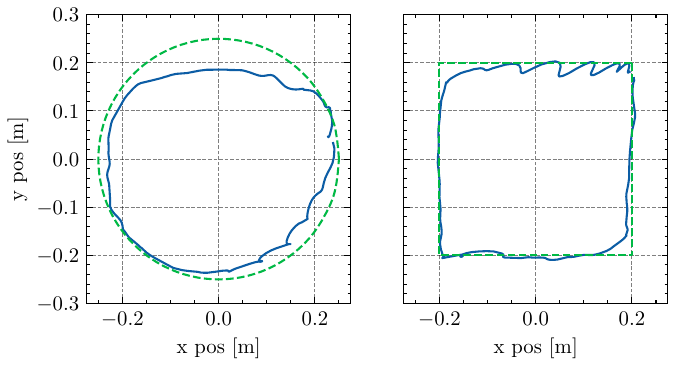}
    \caption{Reference and executed trajectory in the $xy$-plane.}
    \label{fig:soft_arm_traj_xy}
  \end{subfigure}
  \caption{Tracking performance of the control policy on the soft robot arm. The top plot shows the trajectory in the $xy$-plane over time, while the bottom plot shows the trajectory in the $xy$-plane.}
  \label{fig:soft_arm_traj}
\end{figure}

The algorithm converges on the soft arm control task rapidly, reaching satisfactory performance in about 30 episodes.
The loss plots of the training are shown in Figure~\ref{fig:soft_arm_loss}.
After 20 training episodes, the control policy is able to follow a desired trajectory reasonably well.
The tracking performance on a circle and a square test trajectory is shown in Figure~\ref{fig:soft_arm_traj}.
The average tracking error is \SI{2.95}{\centi\meter}.

\revision{
In comparison, previous work on the same hardware by~\citet{Chen24S2C2AFlexible} reports \SI{2.8}{\centi\meter} average error in the position tracking task.
To achieve this, their method involves first training \ac{NN} models of the arm that predict arm state and motor command from each segment's configuration.
Then, trajectory optimization is performed on the state-to-configuration model, and the configuration-space trajectory is tracked with the configuration-to-action model.
While achieving similar accuracy, the proposed online learning algorithm is significantly less complex and does not require specific prior knowledge of the system.
}

\section{Discussion}
\label{sec:discussion}



This work demonstrates an alternative approach to controlling robotic systems without using a simulator.
Instead of relying on a sim-to-real strategy, the presented algorithm learns a control policy directly on the real robot.
The algorithm relies on little prior knowledge, data, or modeling of the system.
In fact, for the presented experiments on the hydraulic excavator and the soft arm, the learning algorithm was initialized in exactly the same way, with the only difference being the dimensions of the state and action spaces, and different normalization factors, as the state and action spaces have very different scales.
By training the control policy solely with real robot data, we gain the additional benefit of not getting control policies biased on a potentially inaccurate simulator, an effect that is often observed in sim-to-real approaches.
During online \revision{training}, if the model is biased, the control policy can exploit the model's errors, collecting new data that updates the model to correct the bias.

The possibility to learn directly on the real robot was not possible due to the sample and computation efficiency of previous approaches.
While model-based \ac{RL} methods are already shown to be more sample efficient than model-free methods, previous approaches like MBPO~\citep{Janner19WhenTrust}, Dreamer~\citep{Wu23DayDreamerWorld}, or TD-MPC~\citep{Hansen24TDMPC2Scalable} require massive rollouts in imagination to perform zeroth-order policy or trajectory optimization.
On the contrary, we combine model-based \ac{RL} with differentiable \ac{RL} techniques, which allows us to perform first-order policy optimization with the learned model.
As a result, we can perform online \revision{training} on the real robot with hours of data while keeping the computational overhead low.

We justify the algorithm design with a theoretical analysis, which interprets the proposed algorithm as two online \revision{optimization} problems.
The difference from standard online learning settings with i.i.d. data and loss gradients suggests the need for the proposed regularization techniques.

\subsection{\revision{Coupling of Model and Policy Learning}}
\begin{figure}
  \centering
  \includegraphics[width=\linewidth]{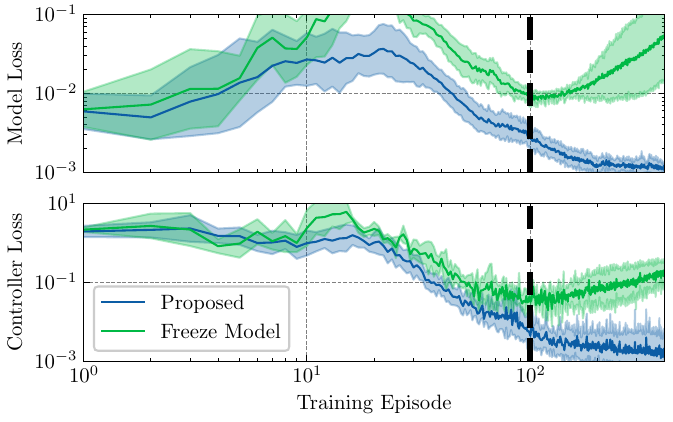}
  \caption{\revision{Comparison between the proposed algorithm and an ablation with the model update paused after 100 episodes. The bold dashed line marks the point where model learning stops.}}
  \label{fig:sim_ablation_pausemodel}
\end{figure}

\begin{figure}
  \centering
  \includegraphics[width=\linewidth]{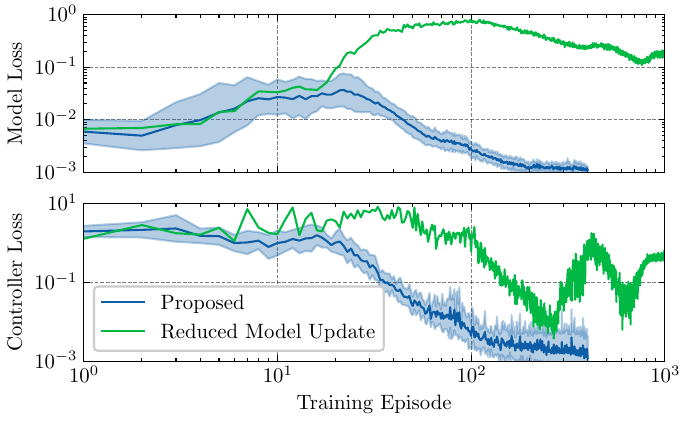}
  \caption{\revision{Comparison between the proposed algorithm and an ablation with the model update at a reduced frequency of every 16 episodes.}}
  \label{fig:sim_ablation_reducedmodelupdate}
\end{figure}

\revision{
In the theoretical interpretation, we treat the model learning and policy learning as two separate online optimization problems and justify the joint convergence with time-scale separation.
When the algorithm is badly tuned and the time-scale separation is not maintained, the coupling between the two learning problems can lead to instability in the learning process.
To demonstrate this effect, we perform an ablation study in the simulation environment of HEAP, where we test two variants of the proposed algorithm.
Figure~\ref{fig:sim_ablation_pausemodel} shows the learning curve of the proposed algorithm with the model update paused after 100 episodes, which leads to a constant model bias in the later part of the learning process.
The figure shows that the control policy performance slowly degrades after the model update is paused, which is a consequence of biased policy gradients due to the model bias.
The model prediction error also increases as the policy slowly diverges from the data distribution on which the model was last updated.
This further worsens the model bias and leads to a vicious cycle of performance degradation.
}

\revision{
Figure~\ref{fig:sim_ablation_reducedmodelupdate} shows the learning curve of the proposed algorithm with the model update at a reduced frequency of every 16 episodes, which shows a cyclic pattern of performance degradation and improvement with prolonged experiments.
In this case, the prediction error stays at a higher level, suggesting the model is constantly unable to explain the data due to the slow update and data distribution shift, which leads to unstable learning of the control policy.
}

\subsection{\revision{Hyperparameter Sensitivity}}
\revision{
For both experiments on the hydraulic excavator and the soft arm, we use the same set of hyperparameters $\alpha=0.01$, $\epsilon=0.05$, and $\eta=0.5$.
While these hyperparameters are found through a grid search in the simulation experiments, we have observed that the algorithm is not very sensitive to the choice of learning rate $\eta$ but is more sensitive to the choice of the regularization parameters $\epsilon$ and $\alpha$.
}

\revision{
We believe this property is a consequence of the preconditioned gradient descent optimization.
As $\Lambda$ is in the form of $\epsilon I + \alpha J J^\top + \nabla \nabla^\top$, the Sherman-Morrison formula suggests
\begin{equation}
  \Lambda^{-1} \nabla = \frac{A^{-1} \nabla}{1 + \nabla^\top A^{-1} \nabla}, \qquad
  A = \epsilon I + \alpha J J^\top.
\end{equation}
The preconditioned gradient update is a scaled version of $A^{-1} \nabla$, where the denominator scales as $\nabla$.
In fact, when $\alpha$ is close to zero and $A^{-1} \approx \nicefrac{1}{\epsilon} I$, the preconditioned step size is bounded by $\nicefrac{1}{2\sqrt{\epsilon}}$, with the maximum achieved when $\|\nabla\| = \sqrt{\epsilon}$.
This implies a "trust region" effect of the regularization term $\epsilon I$.
}

\subsection{\revision{Limitations}}
The presented algorithm is not without limitations.
The theoretical analysis currently assumes that the data distribution shift is bounded.
While this has been true in our experiments, the actual data distribution shift is a consequence of the control policy being updated, assuming time-invariant robot dynamics and a stationary task distribution.
The control policy update, on the other hand, is based on the model that is updated online.
Therefore, the separate discussion of the two components does not consider the full complexity of the problem.
A rigorous analysis that considers the overall stability of such online model learning would be a valuable addition to the theoretical analysis, not only to guarantee the stability of our algorithm, but also to serve as a general framework for analyzing model-based \ac{RL} algorithms as well.
This could be performed by using similar analysis for related problem settings, such as performative prediction~\citep{Hardt25PerformativePrediction, Perdomo20PerformativePrediction} and decision-dependent stochastic optimization~\citep{He25DecisionDependentStochastic}.

In this project, we focused on trajectory-tracking tasks for robots with continuous control.
While this is a common task with many use cases, many complex robotic tasks are described by discrete or sparse reward functions, such as grasping or manipulation tasks.
Many manipulation tasks are also contact-rich, which can lead to discontinuous dynamics.
To apply the proposed algorithm to such tasks, it is necessary to extend the proposed algorithm to use a latent state representation, similar to the implementation of Dreamer~\citep{Hafner19LearningLatent}.
In such complex tasks, the option to train from scratch on the real robot also needs to be reconsidered, as we have not evaluated the proposed algorithm on such tasks.
Different pretraining strategies could be used to initialize the model, the control policy, or both.
Expert demonstration data could also be added to the data buffer to accelerate model learning in the task distribution.

Finally, outliers and noise in the data can have a significant impact on the online learning algorithm.
While learning in a simulation environment might be affected by the bias of the simulator, learning on the real robot is unavoidably affected by noise and outliers in the data.
The noisy data leads to a high-variance gradient estimate for both model and control policy training.
While the issue in model training could be mitigated by using a growing mini-batch size over the accumulating data buffer, the control policy update, which is based on the limited on-policy data, is more sensitive to noise.


\section{Conclusion}
\label{sec:conclusion}


In this paper, we propose a model-based \ac{RL} algorithm for learning to control robots directly on the real system.
The algorithm combines online model learning from common model-based \ac{RL} algorithms with analytical policy gradient updates, which allows for sample-efficient online \revision{training}.
We justify the convergence of the algorithm from the perspective of an online optimization problem, and discuss how the analysis relates to other model-based \ac{RL} and analytical policy gradient algorithms.

We evaluate the proposed algorithm on two distinct robotic systems that have modeling and control challenges: a hydraulic excavator arm and a cable-driven soft robot.
The \revision{proposed} algorithm was deployed on both systems, without any prior knowledge of the system dynamics or control policy, and achieves comparable performance to previously developed methods that rely on a complex workflow involving separate data collection, model learning, and controller design steps.
The results show great potential for the proposed algorithm to be used in real-world robotic applications, by facilitating cross-embodiment transfer of control policies, in particular with robot systems with dynamics that are difficult to model.

Directions for future work include extending the algorithm to more complex tasks, such as manipulation tasks with discrete or sparse rewards, and to robot systems with discontinuous dynamics.
On the theoretical side, the online \revision{optimization} analysis could be extended to consider the coupling between model learning and data distribution shift.

\begin{acks}
The authors would like to thank Mathieu Sandoz, Riccardo Maggioni, and Filippo Spinelli for their help with the experiment on HEAP, Cheng Pan for his help with the experiment on the soft arm, and Pascal Egli and Chenhao Li for their helpful discussions.
\end{acks}

\begin{funding}
  This project has received funding from the European Union's Horizon Europe Framework Programme under grant agreement No 101070405 and from NCCR Automation.
\end{funding}

\begin{sm}
  The supplementary material of this paper can be found at ...
\end{sm}

\bibliographystyle{SageH}
\bibliography{online_learning}

\section{Appendix}
\label{chap:appendix}


\subsection{Derivation of model learning regret}
\label{sec:proof_model_regret_bound}
The model learning part of Algorithm~\ref{alg:online_mbrl} can be analyzed as an online learning problem with shifting data distribution.
While the data distribution, induced by the updating policy, is in fact affected by the learned model, we perform the analysis of the model learning problem itself, isolated from the policy update.
To do this, we could consider the model parameter update as an SGD process on a slowly growing dataset, which has a predefined data distribution sequence unknown to the algorithm.

\begin{figure}[t]
\centering
\includegraphics[width=\linewidth]{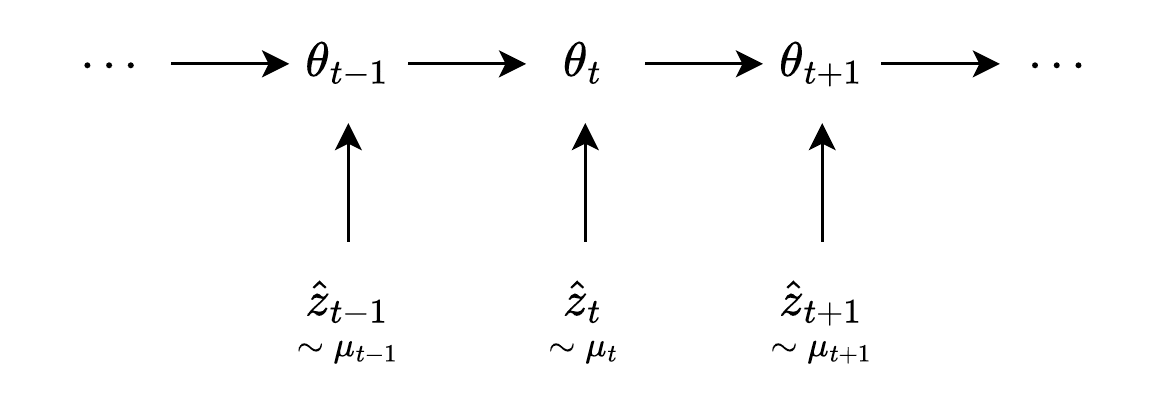}
\caption{SGD update with shifting episode-wise data. Given the sequence $\{\mu_t\}$, draws $\hat z_t\sim\mu_t$ and $\hat z_{t+1}\sim\mu_{t+1}$ are independent conditioning on current parameters.}
\label{fig:model_learning_diagram}
\end{figure}

We first consider a variation of the standard SGD scheme, where the data distribution in each episode shifts.
Let $\{\mu_t\}_{t\ge1}$ be a sequence of data distributions.
In our model learning case, it is defined on the space of state-action-next state triplets on $\mathcal Z = \mathcal X \times \mathcal U \times \mathcal X$.
At episode $t$, a mini-batch element $\hat z_t\sim\mu_t$ is drawn and we perform one SGD step
\begin{equation*}
\theta_t = \theta_{t-1} - \alpha\nabla_\theta \ell(\theta_{t-1};\hat z_t).
\end{equation*}
The dependencies between the random variables $\hat z_t$ and $\theta_{t-1}$ are illustrated in Figure~\ref{fig:model_learning_diagram}.
Due to the resampling of $\hat z_t$ at each episode, there is no dependency between $\hat z_t$ and $\hat z_{t+1}$ conditioning on the model parameter $\theta_t$.

Define the episode-wise optimal parameter w.r.t.\ the current data distribution
\begin{equation*}
\theta_t^{\star}\in\arg\min_{\theta}\ \mathbb E_{z\sim\mu_t}\ell(\theta;z),
\end{equation*}
and the episode-to-episode drift
\begin{equation*}
\widetilde\Delta_t \doteq \|\mu_t-\mu_{t-1}\|_{\mathrm{TV}}\qquad (t\ge2),\quad \widetilde\Delta_1:=0.
\end{equation*}
We study the regret on model prediction loss w.r.t. the current data distribution,
\begin{equation*}
\widetilde R_T \doteq \sum_{t=1}^T \underbrace{\Big(\mathbb E_{z\sim\mu_t}\ell(\theta_t;z)-\mathbb E_{z\sim\mu_t}\ell(\theta_t^\star;z)\Big)}_{=:~\widetilde Y_t},
\end{equation*}

For any $t\ge2$.
We could decompose the optimality gap at episode $t$ as
\begin{align*}
\widetilde Y_t
=& \underbrace{\mathbb E_{z\sim\mu_t}\!\big[\ell(\theta_t;z)\big] - \mathbb E_{z\sim\mu_{t-1}}\!\big[\ell(\theta_t;z)\big]}_{\text{(I)}_t}\\
 &+ \underbrace{\mathbb E_{z\sim\mu_{t-1}}\!\big[\ell(\theta_t;z) - \ell(\theta_{t-1};z)\big]}_{\text{(II)}_t} + \underbrace{\widetilde{Y}_{t-1}}_{\text{(III)}_t}\\
 &+ \underbrace{\mathbb E_{z\sim\mu_{t-1}}\!\big[\ell(\theta_{t-1}^\star;z) - \ell(\theta_t^\star;z)\big]}_{\text{(IV)}_t\le0}\\
 &+ \underbrace{\mathbb E_{z\sim\mu_{t-1}}\!\big[\ell(\theta_t^\star;z)\big]-\mathbb E_{z\sim\mu_t}\!\big[\ell(\theta_t^\star;z)\big]}_{\text{(V)}_t}.
\end{align*}

Since we assume the loss function $\ell$ is Lipschitz continuous on the compact set $\Theta \times \mathcal{Z}$, it attains a finite maximum (Extreme Value Theorem).
Hence there exists
\begin{equation*}
  K \doteq \sup_{\left(\theta, z\right) \in \Theta \times \mathcal{Z}} \left|\ell \left(\theta; z\right)\right| < \infty.
\end{equation*}
Then, the drift pieces satisfy
\begin{equation*}
|\text{(I)}_t| \le 2K\widetilde\Delta_t,\qquad |\text{(V)}_t|\le 2K\widetilde\Delta_t,
\end{equation*}
by bounded loss and the $\mathrm{TV}$ inequality.
For the other terms, we use the update rule
$\theta_t=\theta_{t-1}-\alpha\nabla_\theta\ell(\theta_{t-1};\hat z_t)$ with $\hat z_t\sim\mu_t$,
$\beta_L$-smoothness, the second-moment bound, and PL: 
{\allowdisplaybreaks
\begin{align*}
&\quad \mathbb{E}_{z\sim\mu_t}\ell(\theta_t;z) - \mathbb{E}_{z\sim\mu_t}\ell(\theta_{t-1};z) \\
&\le - \alpha\mathbb{E}_{z, \hat{z}\sim\mu_t}\big[\nabla_\theta\ell(\theta_{t-1};z)^\mathrm{T}\nabla_\theta\ell(\theta_{t-1};\hat z_t)\big] \\ &\quad + \frac{\alpha^2 \beta_L}{2}\mathbb{E}_{\hat{z}\sim\mu_t} \left[\left\|\nabla_\theta\ell(\theta_{t-1};\hat z_t)\right\|^2\right] \\
&= -\alpha\gamma\left(1-\frac{\alpha\beta_L}{2}\right) \mathbb{E}_{z\sim\mu_t} \left[\ell(\theta_{t-1};z) - \ell(\theta_t^\star;z)\right] \\ &\quad + \frac{\alpha^2 \beta_L}{2}\frac{\sigma^2}{B}, %
\end{align*} %
} %
where $\gamma$ is the \ac{PL} constant.
In the last step we use the standard assumption in SGD that the variance of the stochastic gradient is bounded by $\sigma^2/B$ with mini-batch size $B$.

Therefore, we have
\begin{multline*}
\text{(II)}_t+\text{(III)}_t \\
\le -\alpha\gamma\left(1-\frac{\alpha\beta_L}{2}\right)\mathbb E_{\mu_t}\left[\ell(\theta_{t-1};z)-\ell(\theta_t^\star;z)\right] \\
+ \mathbb E_{\mu_{t-1}}\left[\ell(\theta_{t-1};z)-\ell(\theta_t^\star;z)\right] +\frac{\alpha^2\beta_L}{2}\frac{\sigma^2}{B}.
\end{multline*}
Combining all the terms (I) to (V), and notice that $\mathbb E_{\mu_t}\left[\ell(\theta_{t-1};z)-\ell(\theta_t^\star;z)\right] - \mathbb E_{\mu_{t-1}}\left[\ell(\theta_{t-1};z)-\ell(\theta_t^\star;z)\right]$ can be bounded again by TV distance, we obtain a recursion for $\widetilde Y_t$:
\begin{gather*}
\widetilde Y_t \le r\widetilde Y_{t-1} + \frac{\alpha^2\beta_L}{2}\frac{\sigma^2}{B} + 4K\widetilde\Delta_t + 4K(1-r)\widetilde\Delta_t,\\
r \doteq 1-\alpha\gamma\Big(1-\frac{\alpha\beta_L}{2}\Big). 
\end{gather*}
One can choose $\alpha$ small enough such that $\alpha \le 1/\beta_L$ and $0<r<1$.
Then $r\leq 1-\nicefrac{\alpha\gamma}{2}$, and the above recursion becomes
\begin{equation}
\label{eq:tildeY-rec}
\widetilde Y_t \le r\widetilde Y_{t-1} + \frac{\alpha^2\beta_L}{2}\frac{\sigma^2}{B} + 8K\widetilde\Delta_t.
\end{equation}

Unrolling \eqref{eq:tildeY-rec} for $t\ge2$ with $\widetilde Y_1\le C$ gives
\begin{align*}
\widetilde Y_t
&\le 8K\sum_{\tau=2}^t r^{t-\tau}\widetilde\Delta_\tau + r^{t-1}C + \sum_{\tau=2}^t r^{t-\tau}\frac{\alpha^2\beta_L}{2}\frac{\sigma^2}{B}.
\end{align*}
Summing $t=2$ to $T$ and using $\sum_{s\ge0} r^s \le \frac{1}{1-r}$ yields
\begin{equation}
\label{eq:tildeR-bound}
\sum_{t=2}^T \widetilde Y_t \le \frac{16K}{\alpha\gamma}\sum_{\tau=2}^T \widetilde\Delta_\tau + \frac{2C}{\alpha\gamma} + \frac{\beta_L\sigma^2}{B}\cdot \frac{\alpha}{\gamma}T,
\end{equation}
which bounds the total regret $\widetilde{R}_T$.

The bound shown in \eqref{eq:tildeR-bound} consists of three parts.
If there is no distribution change in the learning process, the first term vanishes and the rest can be balanced by choosing $\alpha\propto 1/\sqrt{T}$ to recover the standard $O(\sqrt{T})$ rate.
However, the influence of the first term must be considered when the data distribution is shifting.

In the model learning setting of Algorithm~\ref{alg:online_mbrl}, $\mu_t$ is the data distribution defined by uniformly sampling the buffered data in $\mathcal{D}_t$.
Recall that we use $\Delta_t$ to denote the episode-to-episode drift of collected data distribution $\nu_t$: $\Delta_t=\|\nu_t-\nu_{t-1}\|_{\mathrm{TV}}$ for $t\ge2$ and $\Delta_1=0$.
By the triangle inequality,
{\allowdisplaybreaks
\begin{align*}
\|\nu_t-\mu_t\|_{\mathrm{TV}}
&\le \frac{1}{t}\sum_{i=1}^t \|\nu_t-\nu_i\|_{\mathrm{TV}}
\le \frac{1}{t}\sum_{i=2}^t (i-1)\Delta_i,\\
\|\mu_t-\mu_{t-1}\|_{\mathrm{TV}}
&\le \frac{1}{t}\|\nu_t-\mu_{t-1}\|_{\mathrm{TV}} \\
&\le \frac{1}{t(t-1)}\sum_{i=2}^{t}(i-1)\Delta_i.
\end{align*}}
Then the first term of \eqref{eq:tildeR-bound} can be bounded by an expression of $\Delta_i$:
\begin{equation*}
\sum_{\tau=2}^T \widetilde\Delta_\tau = \sum_{\tau=2}^T \frac{1}{\tau(\tau-1)} \sum_{i=2}^{\tau}(i-1)\Delta_i \leq \sum_{\tau=2}^{T} \Delta_\tau
\end{equation*}

In Algorithm~\ref{alg:online_mbrl}, we only care about the model loss on the latest episode of data, as we only calculate policy gradients based on the most recent data.
Therefore, to analyze the model learning regret on
\begin{multline*}
\bar R_T \doteq \sum_{t=1}^T \Big(\mathbb E_{z\sim\nu_t}\ell(\theta_t;z)-\mathbb E_{z\sim\nu_t}\ell(\theta_t^{\diamond};z)\Big)\\
\theta_t^{\diamond}\in\arg\min_\theta \mathbb E_{z\sim\nu_t}\ell(\theta;z),
\end{multline*}
we decompose it as
\begin{multline*}
\mathbb E_{z\sim\nu_t}\ell(\theta_t;z)-\mathbb E_{z\sim\nu_t}\ell(\theta_t^{\diamond};z)\\
=\underbrace{\mathbb E_{\nu_t}\ell(\theta_t;z)-\mathbb E_{\mu_t}\ell(\theta_t;z)}_{(\mathrm A)_t}
+\underbrace{\mathbb E_{\mu_t}\ell(\theta_t;z)-\mathbb E_{\mu_t}\ell(\theta_t^\star;z)}_{(\mathrm B)_t=\widetilde Y_t}\\
+\underbrace{\mathbb E_{\mu_t}\ell(\theta_t^\star;z) - \mathbb E_{\mu_t}\ell(\theta_t^{\diamond};z)}_{\le 0}
+\underbrace{\mathbb E_{\mu_t}\ell(\theta_t^\diamond;z)-\mathbb E_{\nu_t}\ell(\theta_t^{\diamond};z)}_{(\mathrm C)_t}.
\end{multline*}
We bound the bridge terms again by TV distance and bounded loss:
\begin{equation*}
(\mathrm A)_t\le 2K\|\nu_t-\mu_t\|_{\mathrm{TV}},\qquad
(\mathrm C)_t\le 2K\|\nu_t-\mu_t\|_{\mathrm{TV}}.
\end{equation*}
Summing and plugging in $\Delta_t$ gives
\begin{multline*}
\sum_{t=1}^T \big((\mathrm A)_t+(\mathrm C)_t\big)
\le 4K \sum_{t=1}^T \|\nu_t-\mu_t\|_{\mathrm{TV}}\\
\le 4K \sum_{t=1}^T \frac{1}{t}\sum_{i=2}^t (i-1)\Delta_i.
\end{multline*}

Combining \eqref{eq:tildeR-bound} and the terms $(\mathrm{A})$ and $(\mathrm{C})$ yields
\begin{align*}
\mathbb E[\bar R_T]
&\le 4K \sum_{t=1}^T \frac{1}{t}\sum_{i=2}^t (i-1)\Delta_i + \frac{16K}{\alpha\gamma}\sum_{\tau=2}^T \Delta_{\tau}\\
&\qquad +\ \frac{2C}{\alpha\gamma} + \frac{\beta_L\sigma^2}{B}\cdot \frac{\alpha}{\gamma}T,
\end{align*}
which is the same form as~\eqref{eq:model_learning_regret_bound_with_delta}.




\end{document}